\documentclass[sigconf]{acmart}

\usepackage{amsmath,amsfonts}
\usepackage{algorithm}
\usepackage{graphicx}
\usepackage{textcomp}
\usepackage{makecell}
\usepackage{colortbl}
\usepackage{adjustbox}
\usepackage{color}
\usepackage{xcolor}
\usepackage[caption = false]{subfig}
\usepackage{hyperref}
\usepackage{amsthm}
\usepackage{microtype}      
\usepackage{algpseudocode}  
\usepackage{soul}

\setcopyright{none} 
\acmConference[Anonymous Submission to ACM CCS 2019]{ACM Conference on Computer and Communications Security}{Due 15 May 2019}{London, TBD}
\acmYear{2019}

\begin{document}
\title{Beware the Black-Box: on the Robustness of Recent Defenses to Adversarial Examples} 

\author{Kaleel Mahmood}
\email{kaleel.mahmood@uconn.edu}
\affiliation{%
 \streetaddress{Department of Computer Science and Engineering}
 \institution{University of Connecticut, USA}
 }

\author{Deniz Gurevin}
\affiliation{%
 \streetaddress{Department of Electrical and Computer Engineering}
 \institution{University of Connecticut, USA}
 }

\author{Marten van Dijk}
\affiliation{%
 \institution{CWI, The Netherlands}
 }
 
\author{Phuong Ha Nguyen}
\affiliation{%
 \institution{eBay, USA}
 }

\begin{abstract}
Many defenses have recently been proposed at venues like NIPS, ICML, ICLR and CVPR. These defenses are mainly focused on mitigating white-box attacks. They do not properly examine black-box attacks. In this paper, we expand upon the analysis of these defenses to include adaptive black-box adversaries. Our evaluation is done on nine defenses including Barrage of Random Transforms, ComDefend, Ensemble Diversity, Feature Distillation, The Odds are Odd, Error Correcting Codes, Distribution Classifier Defense, K-Winner Take All and Buffer Zones. Our investigation is done using two black-box adversarial models and six widely studied adversarial attacks for CIFAR-10 and Fashion-MNIST datasets. Our analyses show most recent defenses (7 out of 9) provide only marginal improvements in security ($<25\%$), as compared to undefended networks. For every defense, we also show the relationship between the amount of data the adversary has at their disposal, and the effectiveness of  adaptive black-box attacks. Overall, our results paint a clear picture: defenses need both thorough white-box and black-box analyses to be considered secure. We provide this large scale study and analyses to motivate the field to move towards the development of more robust black-box defenses. 
\end{abstract}



\maketitle
\section{Introduction}\label{sec:intro}

Convolutional Neural Networks (CNNs) are widely used for image classification\cite{KrizhevskySutskeverHinton2012, SimonyanZisserman2015} and object detection. Despite their widespread use, CNNs have been shown to be vulnerable to adversarial examples \cite{goodfellow2014explaining}. Adversarial examples are clean images which have malicious noise added to them. This noise is small enough so that humans can visually recognize the images, but CNNs misclassify them.  

Adversarial examples can be created through white-box or black-box attacks \cite{carlini2019}, depending on the assumed adversarial model. White-box attacks create adversarial examples by directly using information about the trained parameters in a classifier (e.g., the weights of a CNN). Black-box attacks on the other hand, assume an adversarial model where the trained parameters of the classifier are secret or unknown. In black-box attacks, the adversary generates adversarial examples by exploiting other information such as querying the classifier~\cite{PapernotMcDanielGoodfellowEtAl2017,ChenPinYu2017,Chen2019Hop}, or using the original dataset the classifier was trained on~\cite{SzegedyZarembaSutskeverEtAl2013,PapernotMG16,BPDApaper,LiuChenLiuEtAl2016}. We can also further categorize black-box attacks based on whether the attack tries to tailor the adversarial example to specifically overcome the defense (adaptive black-box attacks), or if the attack is fixed regardless of the defense (non-adaptive black-box attacks). In terms of attacks, we focus on adaptive black-box adversaries. A natural question is why do we choose this scope?

1) White-box robustness does not automatically mean black-box robustness. In security communities such as cryptology, black-box attacks are considered strictly weaker than white-box attacks. This means that if a defense is shown to be secure against a white-box adversary, it would also be secure against a black-box adversary. In the field of adversarial machine learning, this principle does NOT always hold true. Why does this happen? In adversarial machine learning, white-box attacks use gradient information directly to create adversarial examples. It is possible to obfuscate this gradient, an effect known as gradient masking~\cite{BPDApaper} and thus make white-box attacks fail. Black-box attacks do not directly use gradient information. As a result, black-box attacks may still be able to work on defenses that have gradient masking. This means adversarial machine learning defenses need to be analyzed under both white-box AND black-box attacks. 

2) White-box adversaries are well studied in most defense papers~\cite{ pang2019improving, verma2019error, jia2019comdefend, raff2019barrage, Xiao2020Enhancing, Kou2020Enhancing, roth2019odds, liu2019feature} as opposed to black-box adversaries. Less attention has been given to black-box attacks, despite the need to test defenses on both types of attacks (as mentioned in our first point). This paper offers a unique perspective by testing defenses under adaptive black-box attacks. By combining the white-box analyses already developed in the literature with the black-box analyses we present here, we give a full security picture.  

Having explained our focus for the type of attacks, we next explain why we chose the following 9 defenses to investigate:

1) Each defense is unique in the following aspect: No two defenses use the exact same set of underlying methods to try and achieve security. We illustrate this point in Table 1. Further in Section~\ref{sec:defsum} we go into specifics about why each individual defense is chosen. As a whole, this diverse group of defenses allows us to evaluate many different competing approaches to security. 

2) Most of the defenses we analyze have been published at NIPS, ICML, ICLR or CVPR. This indicates the machine learning community and reviewers found these approaches worthy of examination and further study.

\subsection{Major Contributions, Related Literature and Paper Organization}
Having briefly introduced the notion of adversarial machine learning attacks and explained the scope of our work, we discuss several other important introductory points. First, we list our major contributions. Second, we discuss literature that is related but distinct from our work. Finally, we give an overview of the organization of the rest of our paper.   

\textbf{Major Contributions:}
\begin{enumerate}
   \item \textit{Comprehensive black-box defense analysis} - Our experiments are comprehensive and rigorous in the following ways: we work with 9 recent defenses and a total of 12 different attacks. Every defense is trained on the same dataset and with the same base CNN architecture whenever possible. Every defense is attacked under the same adversarial model. This allows us to directly compare defense results. It is important to note some papers use different adversarial models which makes comparisons across papers invalid~\cite{carlini2019evaluating}. 
   \item \textit{Adaptive adversarial strength study} - In this paper we are the first (to the best of our knowledge) to show the relationship between each of the 9 defenses and the strength of an adaptive black-box adversary. Specifically, for every defense we are able to show how its security is effected by varying the amount of training data available to an adaptive black-box adversary (i.e., $100\%$, $75\%$, $50\%$, $25\%$ and $1\%$). 
   \item \textit{Open source code and detailed implementations} - One of our main goals of this paper is to help the community develop stronger black-box adversarial defenses. To this end, we publicly provide code for our experiments \href{https://github.com/MetaMain/BewareAdvML}{\textbf{here}}. In addition, in our appendix we give detailed instructions for how we implemented each defense and what experiments we ran to fine tune the hyperparameters of the defense.  
 \end{enumerate}
\textbf{Related Literature:} There are a few works that are related but distinctly different from our paper. We briefly discuss them here. As we previously mentioned, the field of adversarial machine learning has mainly been focused on white-box attacks on defenses. Works that consider white-box attacks on multiple defenses include~\cite{carlini2017adversarial},~\cite{he2017adversarial},~\cite{tramer2020adaptive},~\cite{Dong_2020_CVPR} and~\cite{nguyen2019buzz}.

In~\cite{carlini2017adversarial} the authors test white-box and black-box attacks on defenses proposed in 2017, or earlier. It is important to note, all the defenses in our paper are from 2018 or later. There is no overlap between our work and the work in~\cite{carlini2017adversarial} in terms of defenses studied. In addition, in~\cite{carlini2017adversarial}, while they do consider a black-box attack, it is not adaptive because they do not give the attacker access to the defense training data.

In~\cite{he2017adversarial}, an ensemble is studied by trying to combine multiple weak defenses to form a strong defense. Their work shows that such a combination does not produce a strong defense under a white-box adversary. None of the defenses covered in our paper are used in~\cite{he2017adversarial}. Also~\cite{he2017adversarial} does not consider a black-box adversary like our work.  

In~\cite{Dong_2020_CVPR}, the authors also do a large study on adversarial machine learning attacks and defenses. It is important to note that they do not consider adaptive black-box attacks, as we define them (see Section~\ref{sec:advModel}). They do test defenses on CIFAR-10 like us, but in this case only one defense (ADP~\cite{pang2019improving}) overlaps with our study. To reiterate, the main threat we are concerned with is adaptive black-box attacks which is not covered in~\cite{Dong_2020_CVPR}.

One of the closest studies to us is~\cite{tramer2020adaptive}. In~\cite{tramer2020adaptive} the authors also study adaptive attacks. However, unlike our analyses which use black-box attacks, they assume a white-box adversary. Our paper is a natural progression from~\cite{tramer2020adaptive} in the following sense: If the defenses studied in~\cite{tramer2020adaptive} are broken under an adaptive white-box adversary, could these defenses still be effective under under a weaker adversarial model? In this case, the model in question would be one that disallows white-box access to the defense, i.e., a black-box adversary. Whether these defenses are secure against adaptive black-box adversaries is an open question, and one of the main questions our paper seeks to answer. 

Lastly, adaptive black-box adversaries have also been studied before in~\cite{nguyen2019buzz}. However, they do not consider variable strength adaptive black-box adversaries as we do. We also cover many defenses that are not included in their paper (Error Correcting Codes, Feature Distillation, Distribution Classifier, K-Winner Take All and ComDefend). Finally, the metric we use to compare defenses is fundamentally different from the metric proposed in~\cite{nguyen2019buzz}. They compare results using a metric that balances clean accuracy and security. In this paper, we study the performance of a defense relative to no defense (i.e., a vanilla classifier).   

\textbf{Paper Organization:} Our paper is organized as follows: in Section~\ref{sec:advModel}, we describe the goal of the adversary mathematically, the capabilities given in different adversarial models and the categories of black-box attacks. In Section~\ref{sec:defsum}, we break down the defenses used in this paper in terms of their underlying defense mechanisms. We also explain why each individual defense was selected for analysis in this paper. In Section~\ref{sec:exp}, we discuss the principal experimental results and compare the performances of the defenses. In Section~\ref{sec:indvResults}, we analyze and discuss each defense individually. We also show the relationship between the security of each defense and the strength (amount of training data) available to an adaptive black-box adversary. We offer concluding remarks in Section~\ref{sec:conclusion}. Lastly, full experimental details and defense implementation instructions are given in the appendix.  


\begin{table*}
\centering
\label{key_idea}
\caption{Defenses analyzed in this paper and the corresponding defense mechanisms they employ. For definitions of the each defense mechanism see Section~\ref{sec:defsum}.}
\scalebox{1.0}{
\begin{tabular}{|c|c|c|c|c|c|c|c|c|c|}
\hline 
Defense Mechanism & \makecell{Ensemble \\ Diversity \\ (ADP)}  & \makecell{Error \\ Correcting \\ Codes \\ (ECOC)}  & \makecell{Buffer \\ Zones\\ (BUZz)}  &\makecell{ Com\\Defend}  &\makecell{ Barrage\\ (BaRT)}  &\makecell{ Distribution \\ Classifier \\ (DistC)}  &\makecell{ Feature\\ Distillation \\ (FD)}  &\makecell{ Odds\\ are \\ Odd} &\makecell{ K-Winner \\ (k-WTA)} \\ \hline \hline
Mutiple Models & \checkmark  & \checkmark  & \checkmark  &  &  &  &  &  &  \\ \hline
\makecell{Fixed Input\\ Transformation} &  &  &\checkmark   & \checkmark  &  &  &\checkmark   &  &  \\ \hline
\makecell{Random Input \\ Transformation} &  &  &  & \checkmark  & \checkmark   & \checkmark  &  & \checkmark   &  \\ \hline
\makecell{Adversarial \\ Detection} &  &  & \checkmark   &  &  &  &  & \checkmark   &  \\ \hline
\makecell{Network \\ Retraining} & \checkmark   & \checkmark   &  &  & \checkmark  & \checkmark  &  &  & \checkmark  \\ \hline
\makecell{Architecture \\ Change} &  & \checkmark   &  &  &  & \checkmark  &  &  & \checkmark  \\\hline
\end{tabular}
}
\end{table*}

\section{Attacks}\label{sec:advModel}
\subsection{Attack Setup}
The general setup for an attack in adversarial machine learning can be defined in the following way~\cite{YuanHeZhuEtAl2017}: The adversary is given a trained classifier $F$ which outputs a class label $l$ for a given input $x$ such that $F(x)=l$. In this paper, the classifiers we consider are deep Convolutional Neural Networks (CNN), and the inputs ($x$) are images. The goal of the adversary is to create an adversarial example from the original input $x$ by adding a small noise $\eta$. The adversarial example that is created is a perturbed version of the original input: $x'=x+\eta$. There are two criteria for the attack to be considered successful:

\begin{enumerate}
\item The adversarial example $x'$ must make the classifier produce a certain class label: $F(x')=c$. Here the certain class label $c$ depends on whether the adversary is attempting a targeted, or untargeted type of attack. In a targeted attack $c$ is a specific wrong class label (e.g., a picture of cat MUST be recognized as a dog by the classifier). On the other hand, if the attack is untargted, the only criteria for $c$ is that it must not be the same as the original class label: $c \neq l$ (e.g., as long as a picture of a cat is labeled by the classifier as anything except a cat, the attack is successful).

\item The noise $\eta$ used to create the adversarial image $x'$ must be barely recognizable by humans. This constraint is enforced by limiting the size of perturbation $\eta$ such that the difference between the original input $x$ and the perturbed input $x'$ is less than a certain distance $d$. This distance $d$ is typically measured~\cite{carlini2019evaluating} using the $l_{p}$ norm: $\|x'-x \|_{p} \leq d$ 
\end{enumerate}
In summary, an attack is considered successful if the classifier produces an output label desired by the adversary $F(x')=c$ and the difference between the original input $x$ and the adversarial sample $x'$ is small enough, $\|x'-x \|_{p} \leq d$.  
\subsection{Adversarial Capabilities}\label{sec:advCap}
In this subsection, we go over what information the adversary can use to create adversarial examples.  Specifically, the adversarial model defines what information is available to the attacker to assist them in crafting the perturbation $\eta$. Such information can be broadly grouped into the following categories:
\begin{enumerate}
 \item \textbf{Having knowledge of the trained parameters and architecture of the classifier}. For example, when dealing with CNNs (as is the focus of this paper) knowing the architecture means knowing precisely which type of CNN is used. Example CNN architectures include VGG-16, ResNet56 etc. Knowing the trained parameters for a CNN means the values of the weights and biases of the network (as well as any other trainable parameters) are visible to the attacker~\cite{carlini2019}.
\item \textbf{Query access to the classifier.} If the architecture and trained parameters are kept private, then the next best adversarial capability is having query access to the target model as a black-box. The main concept here is that the adversary can adaptively query the classifier~\cite{brendel2018decisionbased} with different inputs to help create the adversarial perturbation $\eta$. Query access can come in two forms. In the stronger version, when the classifier is queried, the entire probability score vector is returned (i.e., the softmax output from a CNN). Naturally this gives the adversary more information to work with because the confidence in each label is given. In the weaker version, when the classifier is queried, only the final class label is returned (the index of the score vector with the highest value). 

\item \textbf{Having access to (part of the) training or testing data}. In general, for any adversarial machine learning attack, at least one example must be used to start the attack. Hence, every attack requires some input data. However, how much input data the adversary has access to depends on the type of attack (or parameters in the attack). Knowing part or all of the training data used to build the classifier can be especially useful when the architecture and trained parameters of the classifier are not available. This is because the adversary can try to replicate the classifier in the defense, by training their own classifier with the given training data~\cite{PapernotMG16}. 
\end{enumerate}
\subsection{Types of Attacks}
The types of attacks in machine learning can be grouped based on the capabilities the adversary needs to conduct the attack. We described these different capabilities in Section~\ref{sec:advCap}. In this section, we describe the attacks and what capabilities the adversary must have to run them. 

\textbf{White-box attacks:} 
Examples of white-box attacks include the Fast Gradient Sign Method (FGSM)~\cite{goodfellow2014explaining}, Projected Gradient Descent (PGD)~\cite{madry2018towards} and Carlini $\&$ Wagner (C$\&$W)~\cite{CarliniWagner2016} to name a few. They require having knowledge of the trained parameters and architecture of the classifier, as well as query access. In white-box attacks like FGSM and PGD, having access to the classifier’s trained parameters allows the adversary to use a form of backpropagation. By calculating the gradient with respect to the input, the adversarial perturbation $\eta$ can be estimated directly. In some defenses, where directly backpropagating on the classifier may not be applicable or yield poor results, it is possible to create attacks tailored to the defense that are more effective. These are referred to as adaptive attacks~\cite{tramer2020adaptive}.
In general, white-box attacks and defenses against them have been heavily focused on in the literature. In this paper, our focus is on black-box attacks. Hence, we only give a brief summary of the white-box attacks as mentioned above. 

\textbf{Black-Box Attacks:} The biggest difference between white-box and black-box attacks is that black-box attacks lack access to the trained parameters and architecture of the defense. As a result, they need to either have training data to build a synthetic model, or use a large number of queries to create an adversarial example. Based on these distinctions, we can categorize black-box attacks as follows:
\begin{enumerate}
\item Query only black-box attacks~\cite{brendel2018decisionbased}. The attacker has query access to the classifier. In these attacks, the adversary does not build any synthetic model to generate adversarial examples or make use of training data. Query only black-box attacks can further be divided into two categories: score based black-box attacks and decision based black-box attacks.
\begin{itemize}
\item Score based black-box attacks. These are also referred to as zeroth order optimization based black-box attacks~\cite{ChenPinYu2017}. In this attack, the adversary adaptively queries the classifier with variations of an input $x$ and receives the output from the softmax layer of the classifier $f(x)$. Using $x, f(x)$ the adversary attempts to approximate the gradient of the classifier $\nabla f$ and create an adversarial example. SimBA is an example of one of the more recently proposed score based black-box attacks~\cite{guo2019simple}. 
\item Decision based black-box attacks. The main concept in decision based attacks is to find the boundary between classes using only the hard label from the classifier. In these types of attacks, the adversary does not have access to the output from the softmax layer (they do not know the probability vector). Adversarial examples in these attacks are created by estimating the gradient of the classifier by querying using a binary search methodology. Some recent decision based black-box attacks include HopSkipJump~\cite{Chen2019Hop} and RayS~\cite{chen2020rays}.
\end{itemize}
\item Model black-box attacks. In model black-box attacks, the adversary has access to part or all of the training data used to train the classifier in the defense. The main idea here is that the adversary can build their own classifier using the training data, which is called the synthetic model. Once the synthetic model is trained, the adversary can run any number of white-box attacks (e.g., FGSM~\cite{goodfellow2014explaining}, BIM~\cite{kurakin2017adversarial}, MIM~\cite{dong2018boosting}, PGD~\cite{madry2018towards}, C$\&$W~\cite{CarliniWagner2016} and EAD~\cite{chen2018ead}) on the synthetic model to create adversarial examples. The attacker then submits these adversarial examples to the defense. Ideally, adversarial examples that succeed in fooling the synthetic model will also fool the classifier in the defense. Model black-box attacks can further be categorized based on how the training data in the attack is used: 
\begin{itemize}
\item Adaptive model black-box attacks~\cite{PapernotMcDanielGoodfellowEtAl2017}. In this type of attack, the adversary attempts to adapt to the defense by training the synthetic model in a specialized way. Normally, a model is trained with dataset $X$ and corresponding class labels $Y$. In an adaptive black-box attack, the original labels $Y$ are discarded. The training data $X$ is re-labeled by querying the classifier in the defense to obtain class labels $\hat{Y}$. The synthetic model is then trained on $(X,\hat{Y})$ before being used to generate adversarial examples. The main concept here is that by training the synthetic model with $(X,\hat{Y})$, it will more closely match or adapt to the classifier in the defense. If the two classifiers closely match, then there will (hopefully) be a higher percentage of adversarial examples generated from the synthetic model that fool the classifier in the defense. To run adaptive black-box attacks, access to at least part of the training data and query access to the defense is required. If only a small percentage of the training data is known (e.g., not enough training data to train a CNN), the adversary can also generate synthetic data and label it using query access to the defense~\cite{PapernotMcDanielGoodfellowEtAl2017}.     
\item Pure black-box attacks~\cite{SzegedyZarembaSutskeverEtAl2013,PapernotMG16,BPDApaper,LiuChenLiuEtAl2016}. In this type of attack, the adversary also trains a synthetic model. However, the adversary does not have query access to make the attack adaptive. As a result, the synthetic model is trained on the original dataset and original labels $(X,Y)$. In essence this attack is defense agnostic (the training of the synthetic model does not change for different defenses). 
\end{itemize}
\end{enumerate}

\begin{table*}[]
\centering
\label{attackTable}
\caption{Adversarial machine learning attacks and the adversarial capabilities required to execute the attack. For a full description of these capabilities, see Section~\ref{sec:advCap}.}
\begin{tabular}{lccll}
 & \multicolumn{4}{c}{Adversarial Capabilities} \\ \cline{2-5} 
\multicolumn{1}{l|}{} & \multicolumn{1}{l|}{Training/Testing Data} & \multicolumn{1}{l|}{Hard Label Query Access} & \multicolumn{1}{l|}{Score Based Query Access} & \multicolumn{1}{l|}{Trained Parameters} \\ \hline
\multicolumn{1}{|l|}{White-Box} & \multicolumn{1}{c|}{} & \multicolumn{1}{c|}{ \checkmark} & \multicolumn{1}{c|}{\checkmark} & \multicolumn{1}{c|}{\checkmark} \\ \hline
\multicolumn{1}{|l|}{Score Based Black-Box} & \multicolumn{1}{c|}{} & \multicolumn{1}{c|}{\checkmark} & \multicolumn{1}{c|}{\checkmark} & \multicolumn{1}{l|}{} \\ \hline
\multicolumn{1}{|l|}{Decision Based Black-Box} & \multicolumn{1}{c|}{} & \multicolumn{1}{c|}{\checkmark} & \multicolumn{1}{l|}{} & \multicolumn{1}{l|}{} \\ \hline
\multicolumn{1}{|l|}{Mixed Black-Box} & \multicolumn{1}{c|}{\checkmark} & \multicolumn{1}{c|}{\checkmark} & \multicolumn{1}{l|}{} & \multicolumn{1}{l|}{} \\ \hline
\multicolumn{1}{|l|}{Pure Black-Box} & \multicolumn{1}{c|}{\checkmark} & \multicolumn{1}{l|}{} & \multicolumn{1}{l|}{} & \multicolumn{1}{l|}{} \\ \hline
\end{tabular}
\end{table*}

\subsection{Our black-box attack scope}

We focus on black-box attacks, specifically the adaptive black-box and pure black-box attacks. Why do we refine our scope in this way? First of all we don’t focus on white-box attacks as mentioned in Section~\ref{sec:intro} as this is well documented in the current literature. In addition, simply showing white-box security is not enough in adversarial machine learning. Due to gradient masking~\cite{BPDApaper}, there is a need to demonstrate both white-box and black-box robustness. When considering black-box attacks, as we explained in the previous subsection, there are query only black-box attacks and model black-box attacks. Score based query black-box attacks can be neutralized by a form of gradient masking~\cite{carlini2019evaluating}. Furthermore, it has been noted that a decision based query black-box attack represents a more practical adversarial model~\cite{chen2019hopskipjumpattack}. However, even with these more practical attacks there are disadvantages. It has been claimed that decision based black-box attacks may perform poorly on randomized models~\cite{carlini2019evaluating, Dong_2020_CVPR}. It has also been shown that even adding a small Gaussian noise to the input may be enough to deter query black-box attacks~\cite{anonymous2021small}. Due to their poor performance in the presence of even small randomization, we do not consider query black-box attacks. 

Focusing on black-box adversaries and discounting query black-box attacks, leaves model black-box attacks. In our analyses, we first use the pure black-box attack because this attack has no adaptation and no knowledge of the defense. In essence it is the least capable adversary. It may seem counter-intuitive to start with a weak adversarial model. However, by using a relatively weak attack we can see the security of the defense under idealized circumstances. This represents a kind of best-case defense scenario. 

The second type of attack we focus on is the adaptive black-box attack. This is the strongest model black-box type of attack in terms of the powers given to the adversary. In our study on this attack, we also vary its strength by giving the adversary different amounts of the original training data ($1\%$, $25\%$, $50\%$, $75\%$ and $100\%$). For the defense, this represents a stronger adversary, one that has query access, training data and an adaptive way to try and tailor the attack to break the defense. In short, we chose to focus on the pure and adaptive black-box attacks. We do this because they do not suffer from the limitations of the query black-box attacks, and they can be used as an efficient and nearly universally applicable security test.

\section{Defense summaries, metrics and datasets}
\label{sec:defsum}
In this paper we investigate 9 recent defenses, Barrage of Random Transforms (BaRT)~\cite{raff2019barrage}, End-to-End Image Compression Models (ComDefend)~\cite{jia2019comdefend}, The Odds are Odd (Odds)~\cite{roth2019odds}, Feature Distillation (FD)~\cite{liu2019feature}, Buffer Zones (BUZz)~\cite{nguyen2019buzz}, Ensemble Diversity (ADP)~\cite{pang2019improving}, Distribution Classifier (DistC)~\cite{Kou2020Enhancing}, Error Correcting Output Codes (ECOC)~\cite{verma2019error} and K-Winner-Take-All (k-WTA)~\cite{Xiao2020Enhancing}.    

In Table 1, we decompose these defenses into the underlying methods they use to try to achieve security. This is by no means the only way these defenses can be categorized and the definitions here are not absolute. We merely provide this hierarchy to provide a basic overview and show common defense themes. The defense themes are categorized as follows: 

\begin{enumerate}
  \item\textbf{ Multiple models} - The defense uses multiple classifiers' for prediction. The classifiers outputs may be combined through averaging (i.e., ADP), majority voting (BUZz) or other methods (ECOC).
  
  \item \textbf{Fixed input transformation} - A non-randomized transformation is applied to the input before classification. Examples of this include, image denoising using an autoencoder (Comdefend), JPEG compression (FD) or resizing and adding (BUZz). 
  
  \item \textbf{Random input transformation} - A random transformation is applied to the input before classification. For example both BaRT and DistC randomly select from multiple different image transformations to apply at run time. 
  
  \item \textbf{Adversarial detection} - The defense outputs a null label if the sample is considered to be adversarially manipulated. Both BUZz and Odds employ adversarial detection mechanisms. 
  \item \textbf{Network retraining} - The network is retrained to accommodate the implemented defense. For example BaRT and BUZz require network retraining to achieve acceptable clean accuracy. This is due to the significant transformations both defenses apply to the input. On the other hand, different architectures mandate the need for network retraining like in the case of ECOC, DistC and k-WTA. Note network retraining is different from adversarial training. In the case of adversarial training, it is a fundamentally different technique in the sense that it can be combined with almost every defense we study. Our interest however is not to make each defense as strong as possible. Our aim is to understand how much each defense improves security on its own. Adding in techniques beyond what the original defense focuses on is essentially adding in confounding variables. It then becomes even more difficult to determine from where security may arise. As a result, we limit the scope of our defenses to only consider retraining when required and do not consider adversarial training. 
  \item \textbf{Architecture change} - A change in the architecture which is made solely for the purposes of security. For example k-WTA uses different activation functions in the convolutional layers of a CNN. ECOC uses a different activation function on the output of the network.
\end{enumerate}
\subsection{Barrage of random transforms}
Barrage of Random Transforms (BaRT)~\cite{raff2019barrage} is a defense based on applying image transformations before classification. The defense works by randomly selecting a set of transformations and a random order in which the image transformations are applied. In addition, the parameters for each transformation are also randomly selected at run time to further enhance the entropy of the defense. Broadly speaking, there are 10 different image transformation groups: JPEG compression, image swirling, noise injection, Fourier transform perturbations, zooming, color space changes, histogram equalization, grayscale transformations and denoising operations. 

\textbf{Prior security studies:} In terms of white-box analyses, the original BaRT paper tests PGD and FGSM. They also test a combined white-box attack designed to deal with randomization. This combinational white-box attack is composed of  expectation over transformation~\cite{athalye2018synthesizing} and  backward pass differentiable approximation~\cite{BPDApaper}. No analysis of the BaRT defense with black-box adversaries is done.    

\textbf{Why we selected it:} In~\cite{carlini2019evaluating}, they claim gradient free attacks (i.e., black-box attacks) most commonly fail due to randomization. Therefore BaRT is a natural candidate to test for black-box security. Also in the original paper, BaRT is only tested with ImageNet. We wanted to see if this defense could be expanded to work on other datasets.

\subsection{End-to-end image compression models}

ComDefend~\cite{jia2019comdefend} is a defense where image compression/reconstruction is done using convolutional autoencoders before classification. ComDefend consists of two modules: a compression convolutional neural network (ComCNN) and a reconstruction convolutional neural network (RecCNN). The compression network transforms the input image into a compact representation by compressing the original 24 bit pixels into compact 12 bit representations. Gaussian noise is then added to the compact representation. Decompression is then done using the reconstruction network and the final output is fed to the classifier. In this defense, retraining of the classifier on reconstructed input data is not required. 

\textbf{Prior security studies:} White-box attacks such as FGSM, BIM and C$\&$W are run on ComDefend. They also vary their threat model between using the $l_{\infty}$ norm and $l_{2}$ norm to create white-box adversarial examples that have different constraints. No black-box attacks are ever presented for the defense.   

\textbf{Why we selected it:} Other autoencoder defenses have fared poorly~\cite{carlini2017magnet}. It is worth studying new autoencoder defenses to see if they work, or if they face the same vulnerabilities as older defense designs. Since ComDefend~\cite{jia2019comdefend} does not study black-box adversaries, our analysis also provides new insight on this defense. 

\subsection{The odds are odd}

The Odds are Odd~\cite{roth2019odds} is a defense based on a statistical test. This test is motivated by the following observation: the behaviors of benign and adversarial examples are different at the logits layer (i.e., the input to the softmax layer). The test works as follows: for a given input image, multiple copies are created and a random noise is added to each copy. This creates multiple random noisy images. The defense calculates the logits values of each noisy image and uses them as the input for the statistical test.

\textbf{Prior security studies:} In the original Odds paper, the statistical test is done in conjunction with adversarial examples generated using PGD (a white-box attack). Further white-box attacks on the Odds were done in~\cite{tramer2020adaptive}. The authors in~\cite{tramer2020adaptive} use PGD and a custom objective function to show the flaws in the statistical test under white-box adversarial model. To the best of our knowledge, no work has been done on the black-box security of the Odds defense.  



\textbf{Why we selected it:} In ~\cite{tramer2020adaptive}, they mention that Odds is based on the common misconception that building a test for certain adversarial examples will then work for all adversarial examples. However, in the black-box setting this still brings up an interesting question: if the attacker is unaware of the type of test, can they still adaptively query the defense and come up with adversarial examples that circumvent the test?

\subsection{Feature distillation}

Feature Distillation (FD) implements a unique JPEG compression and decompression technique to defend against adversarial examples. Standard JPEG compression/decompression preserves low frequency components. However, it is claimed in~\cite{liu2019feature} that CNNs learn features which are based on high frequency components. Therefore, the authors propose a compression technique where a smaller quantization step is used for CNN accuracy-sensitive frequencies and a larger quantization step is used for the remaining frequencies. The goal of this technique is two-fold. First, by maintaining high frequency components, the defense aims to preserve clean accuracy. Second, by reducing the other frequencies, the defense tries to eliminate the noise that make samples adversarial. Note this defense does have some parameters which need to be selected through experimentation. For the sake of brevity, we provide the experiments for selecting these parameters in the appendix. 

\textbf{Prior security studies:} In the original FD paper, the authors test their defense against standard white-box attacks like FGSM, BIM and C$\&$W. They also analyze their defense against the backward pass differentiable approximation~\cite{BPDApaper} white-box attack. In terms of black-box adversaries, they do test a very simple black-box attack. In this attack,  samples are generated by first training a substitute model. However, this black-box adversary cannot query the defense to label its training data, making it extremely limited. Under our attack definitions, this is not an adaptive black-box attack. 



\textbf{Why we selected it:} A common defense theme is the utilization of multiple image transformations like in the case of BaRT, BUZz and DistC. However, this requires a cost in the form of network retraining and/or clean accuracy. If a defense could use only one type of transformation (as done in FD), it may be possible to significantly reduce those costs. To the best of our knowledge, so far no single image transformation has accomplished this, which makes the investigation of FD interesting.

\subsection{Buffer zones}

Buffer Zones (BUZz) employs a combination of techniques to try and achieve security. The defense is based on unanimous majority voting using multiple classifiers. Each classifier applies a different fixed secret transformation to its input. If the classifiers are unable to agree on a class label, the defense marks the input as adversarial. The authors also note that a large drop in clean accuracy is incurred due to the number of defense techniques employed.

\textbf{Prior security studies:} BUZz is the only defense on our list that experiments with a similar black-box adversary (one that has access to the training data and can query the defense). However, as we explain below, their study has room to further be expanded upon.  

\textbf{Why we selected it:} We selected this defense to study because it specifically claims to deal with the exact adversarial model (adaptive black-box) that we work with. However, in their paper they only use a single strength adversary (i.e., one that uses the entire training dataset). We test across multiple strength adversaries (see Section~\ref{sec:indvResults}) to see how well their defense holds up.

\subsection{Improving adversarial robustness via promoting ensemble diversity}

Constructing ensembles of enhanced networks is one defense strategy to improve the adversarial robustness of classifiers. However, in an ensemble model, the lack of interaction among individual members may cause them to return similar predictions. This defense proposes a new notion of ensemble diversity by promoting the diversity among the predictions returned by members of an ensemble model using an adaptive diversity promoting (ADP) regularizer, which works with a logarithm of ensemble diversity term and an ensemble entropy term~\cite{pang2019improving}. The ADP regularizer helps non-maximal predictions of each ensemble member to be mutually orthogonal, while the maximal prediction is still consistent with the correct label. This defense employs a different training procedure where the ADP regularizer is used as the penalty term and the ensemble network is trained interactively. 

\textbf{Prior security studies:} ADP has widely been studied in the context of white-box  security in~\cite{pang2019improving},~\cite{tramer2020adaptive} and~\cite{Dong_2020_CVPR}. In the original paper in which ADP was proposed, they tested the defense against white-box attacks like FGSM, BIM, PGD, C$\&$W and EAD. In~\cite{tramer2020adaptive}, they use different attack parameters (more iterations) in order to show the defense was not as robust as previously thought. These results are further supported by white-box attacks done on ADP and reported in~\cite{Dong_2020_CVPR}.They use FGSM, BIM and MIM (as well as others white-box attacks) in~\cite{Dong_2020_CVPR} to further analyze the robustness of ADP. They also test some black-box attacks on ADP in~\cite{Dong_2020_CVPR}, but these attacks are transfer based and boundary based. They do not test our adaptive type of black-box attack in~\cite{Dong_2020_CVPR}.  

\textbf{Why we selected it:} It has been shown that adversarial samples can have high transferability~\cite{PapernotMcDanielGoodfellowEtAl2017}. Model black-box attacks  have a basic underlying assumption: adversarial samples that fool the synthetic model will also fool the defense. ADP trains networks to specifically enhance diversity which could mitigate the transferability phenomena. If the adversarial transferability between networks is indeed really mitigated, then black-box attacks should not be effective.

\subsection{Enhancing transformation-based defenses against adversarial attacks with a distribution classifier}

The basic idea of this defense is that if the input is adversarial, basing the predicted class on the softmax output may yield a wrong result. Instead in this defense the input is randomly transformed multiple times, to create many different inputs. Each transformed input yields a softmax output from the classifier. Prediction is then done on the distribution of the softmax outputs~\cite{Kou2020Enhancing}. To classify the softmax distributions, a separate distributional classifier is trained. 

\textbf{Prior security studies:} In~\cite{Kou2020Enhancing}, white-box attacks on the defense were done using methods like FGSM, IFGSM and C$\&$W. Query only black-box attacks were also studied, but by our definition, no adaptive black-box attacks were ever considered for this defense.  

\textbf{Why we selected it:} In~\cite{Kou2020Enhancing}, the defense is tested with query only black-box attacks as we previously mentioned. However, it does not test any model black-box attacks. This defense is built on~\cite{XieWangZhangEtAl2017} which was initially a promising randomization defense that was broken in~\cite{BPDApaper}. Whether the combination of a new classification scheme and randomization can achieve model black-box security is an open question.



\subsection{Error correcting output codes }

The Error Correcting Output Codes (ECOC)~\cite{verma2019error} defense uses the idea of coding theory and changes the output representation in a network to codewords. There are three main ideas of the defense. First, is the use of a special sigmoid decoding activation function instead of the softmax function. This function allocates the non-trivial volume in logit space to uncertainty. This makes the attack surface smaller to the attacker who tries to craft adversarial examples. Second, a larger Hamming distance between the codewords is used to increase the distance between two high-probability regions for a class in logit space. This forces the adversary to use larger perturbations in order to succeed. Lastly, the correlation between outputs is reduced by training an ensemble model.

\textbf{Prior security studies:} In~\cite{verma2019error}, the authors test ECOC against white-box attacks like PGD and C$\&$W. A further white-box analysis of ECOC is done in~\cite{tramer2020adaptive}, where PGD with a custom loss function is used. Through this modified PGD, the authors in~\cite{tramer2020adaptive} are able to significantly reduce the robustness of the ECOC defense in the white-box setting. No black-box analyses of ECOC are ever considered in~\cite{tramer2020adaptive} or~\cite{verma2019error}.

\textbf{Why we selected it:} Much like ADP, this method relies on an ensemble of models. However unlike ADP, this defense is based on coding theory and the original paper does not consider a black-box adversary. The authors in~\cite{tramer2020adaptive} were only able to come up with an effective attack on ECOC in the white-box setting. Therefore, exploring the black-box security of this defense is of interest. 


\subsection{k-winner-take-all}

In k-Winner-Take-All (k-WTA)~\cite{Xiao2020Enhancing} a special activation function is used that is $C^{0}$ discontinuous. This activation function mitigates white-box attacks through gradient masking. The authors claim this architecture change is nearly free in terms of the drop in clean accuracy. 

\textbf{Prior security studies:} In the original k-WTA paper~\cite{Xiao2020Enhancing} the authors test their defense against white-box attacks like PGD, MIM and C$\&$W. They also test against a weak transfer based  black-box attack that is not adaptive. They do not consider a black-box adversary that has access to the entire training dataset and query access like we assume in our adversarial model. Further white-box attacks against k-WTA were done in~\cite{tramer2020adaptive}. The authors in~\cite{tramer2020adaptive} used PGD with more iterations (400) and also considered a special averaging technique to better estimate the gradient of the network.

\textbf{Why we selected it:} The authors of the defense claim that k-WTA performs better under model black-box attacks than networks that use ReLU activation functions. If this claim is true, this would be the first defense in which gradient masking could mitigate both white-box and black-box attacks. In~\cite{tramer2020adaptive}, they already showed the vulnerability of this defense to white-box attacks. Additionally, in~\cite{tramer2020adaptive} they hypothesize a black-box adversary that queries the network may work well against this defense, but do not follow up with any experiments. Therefore, this indicates k-WTA still lacks proper black-box security experiments and analyses.


 
\subsection{Defense metric}
\label{sec:defmet}
In this paper, our goal is to demonstrate what kind of gain in security can be achieved by using each defense against a black-box adversary. Our aim is not to claim any defense is broken. To measure the improvement in security, we use a simple metric: \textit{Defense accuracy improvement}.

Defense accuracy improvement is the percent increase in correctly recognized adversarial examples gained when implementing the defense as compared to having no defense. The formula for defense accuracy improvement for the $i^{th}$ defense is defined as:

\begin{equation}
    A_{i} = D_{i} - V
\end{equation}

We compute the defense accuracy improvement $A_{i}$ by first conducting a specific black-box attack on a vanilla network (no defense). This gives us a vanilla defense accuracy score $V$. The vanilla defense accuracy is the percent of adversarial examples the vanilla network correctly identifies. We run the same attack on a given defense. For the $i^{th}$ defense, we will obtain a defense accuracy score of $D_{i}$. By subtracting $V$ from $D_{i}$ we essentially measure how much security the defense provides as compared to not having any defense on the classifier. 

For example if $V \approx 99\%$, then the defense accuracy improvement $A_{i}$ can be $0$, but at the very least should not be negative. If $V\approx 85\%$, then a defense accuracy improvement of $10\%$ may be considered good. If $V\approx 40\%$, then  
we want at least a $25\%$ defense accuracy improvement, for the defense to be considered effective (i.e. the attack fails more than half of the time when the defense is implemented). While sometimes an improvement is not possible (e.g. when $V \approx 99\%$) there are many cases where attacks works well on the undefended network and hence there are places where large improvements can be made. Note to make these comparisons as precise as possible, almost every defense is built with the same CNN architecture. Exceptions to this occur in some cases, which we fully explain in the appendix.

\subsection{Datasets}

In this paper, we test the defenses using two distinct datasets, CIFAR-10~\cite{CIFAR10ref} and Fashion-MNIST~\cite{Han2017FashionMNIST}. CIFAR-10 is a dataset comprised of 50,000 training images and 10,000 testing images. Each image is 32x32x3 (a 32x32 color image) and belongs to 1 of 10 classes. The 10 classes in CIFAR-10 are airplane, car, bird, cat, deer, dog, frog, horse, ship and truck. Fashion-MNIST is a 10 class dataset with 60,000 training images and 10,000 test images. Each image in Fashion-MNIST is 28x28 (grayscale image). The classes in Fashion-MNIST correspond to t-shirt, trouser, pullover, dress, coat, sandal, shirt, sneaker, bag and ankle boot.

\textbf{Why we selected them:} We chose the CIFAR-10 defense because many of the existing defenses had already been configured with this dataset. Those defenses already configured for CIFAR-10 include ComDefend, Odds, BUZz, ADP, ECOC, the distribution classifier defense and k-WTA. We also chose CIFAR-10 because it is a fundamentally challenging dataset. CNN configurations like ResNet do not often achieve above $94\%$ accuracy on this dataset~\cite{he2016deep}. In a similar vein, defenses often incur a large drop in clean accuracy on CIFAR-10 (which we will see later in our experiments with BUZz and BaRT for example). This is because the amount of pixels that can be manipulated without hurting classification accuracy is limited. For CIFAR-10, each image only has in total 1,024 pixels. This is relatively small when compared to a dataset like ImageNet~\cite{deng2009imagenet}, where images are usually 224x224x3 for a total of 50,176 pixels (49 times more pixels than CIFAR-10 images). In short, we chose CIFAR-10 as it is a challenging dataset for adversarial machine learning and many of the defenses we test were already configured with this dataset in mind. 

For Fashion-MNIST, we primarily chose it for two main reasons. First, we wanted to avoid a trivial dataset on which all defenses might perform well. For example, CNNs can already achieve a clean accuracy of $99.7\%$ on a dataset like MNIST~\cite{Han2017FashionMNIST}. Testing on such types of datasets would not work towards the main aim of our paper, which is to distinguish defenses that perform significantly better in terms of security and clean accuracy. The second reason we chose Fashion-MNIST is for its differences from CIFAR-10. Specifically, Fashion-MNIST is a non-color dataset and contains very different types of images than CIFAR-10. In addition, many of the defenses we tested were not originally designed for Fashion-MNIST. This brings up an interesting question, can previously proposed defenses be readily adapted to work with different datasets. To summarize, we chose Fashion-MNIST for its difficult to learn and its differences from CIFAR-10.  
\section{Principal Experimental Results}
\label{sec:exp}

In this section, we conduct experiments to test the black-box security of the 9 defenses. We measure the results using the metric defense accuracy improvement (see Section~\ref{sec:defmet}). For each defense, we test it under a pure black-box adversary, and five different strength adaptive black-box adversaries. The strength of the adaptive black-box adversary is determined by how much of the original training dataset they are given access to (either $100\%$, $75\%$, $50\%$, $25\%$ or $1\%$). For every adversary, once the synthetic model is trained, we use 6 different methods (FGSM~\cite{goodfellow2014explaining}, BIM~\cite{kurakin2017adversarial}, MIM~\cite{dong2018boosting}, PGD~\cite{madry2018towards}, C$\&$W~\cite{CarliniWagner2016} and EAD~\cite{chen2018ead}) to generate adversarial examples. 

Before going into a thorough analysis of our results, we briefly introduce the figures and tables that show our experimental results. Figure~\ref{fig:cifar10Mixed} and Figure~\ref{fig:pureCIFAR10} illustrate the defense accuracy improvement of all the defenses under a $100\%$ strength adaptive black-box adversary (Figures~\ref{fig:cifar10Mixed}) and a pure black-box adversary (Figures~\ref{fig:pureCIFAR10}) for the CIFAR-10 dataset. Likewise, for Fashion-MNIST, Figure~\ref{fig:fmnistMixed} shows the defense accuracy improvement under a $100\%$ strength adaptive black-box adversary and Figure~\ref{fig:fMNISTPure} shows the defense accuracy improvement under a pure black-box adversary. For each of these figures, we report the vanilla accuracy numbers in a chart below the graph. Figures~\ref{fig:bartadaptiveVar} through~\ref{fig:kWinneradaptiveVar} show the relationship between the defense accuracy and the strength of the adversary (how much training data the adversary has access to). Figures~\ref{fig:bartadaptiveVar} through~\ref{fig:kWinneradaptiveVar} show this relationship for each defense, on both CIFAR-10 and Fashion-MNIST. The corresponding values for the figures are given in Tables~\ref{table:CIFARPure} through~\ref{table:FM100}.

Considering the range of our experiments (9 defenses, 6 adversarial models, 6 methods to generate adversarial samples and 2 datasets), it is infeasible to report all the results and experimental details in just one section. Instead, we organize our experimental analysis as follows. In this section, we present the most pertinent results in Figures~\ref{fig:cifar10Mixed} and~\ref{fig:fmnistMixed} and  give the principal takeaways. For readers interested in a specific defense or attack results, in Section~\ref{sec:indvResults} we give a comprehensive break down of the results for each defense, dataset and attack. For anyone wishing to recreate our experimental results, we give complete implementation details for every attack and defense in the appendix.  

\subsection{Principal results}

\begin{figure*}[!htb]
\centering
\includegraphics[scale=0.8]{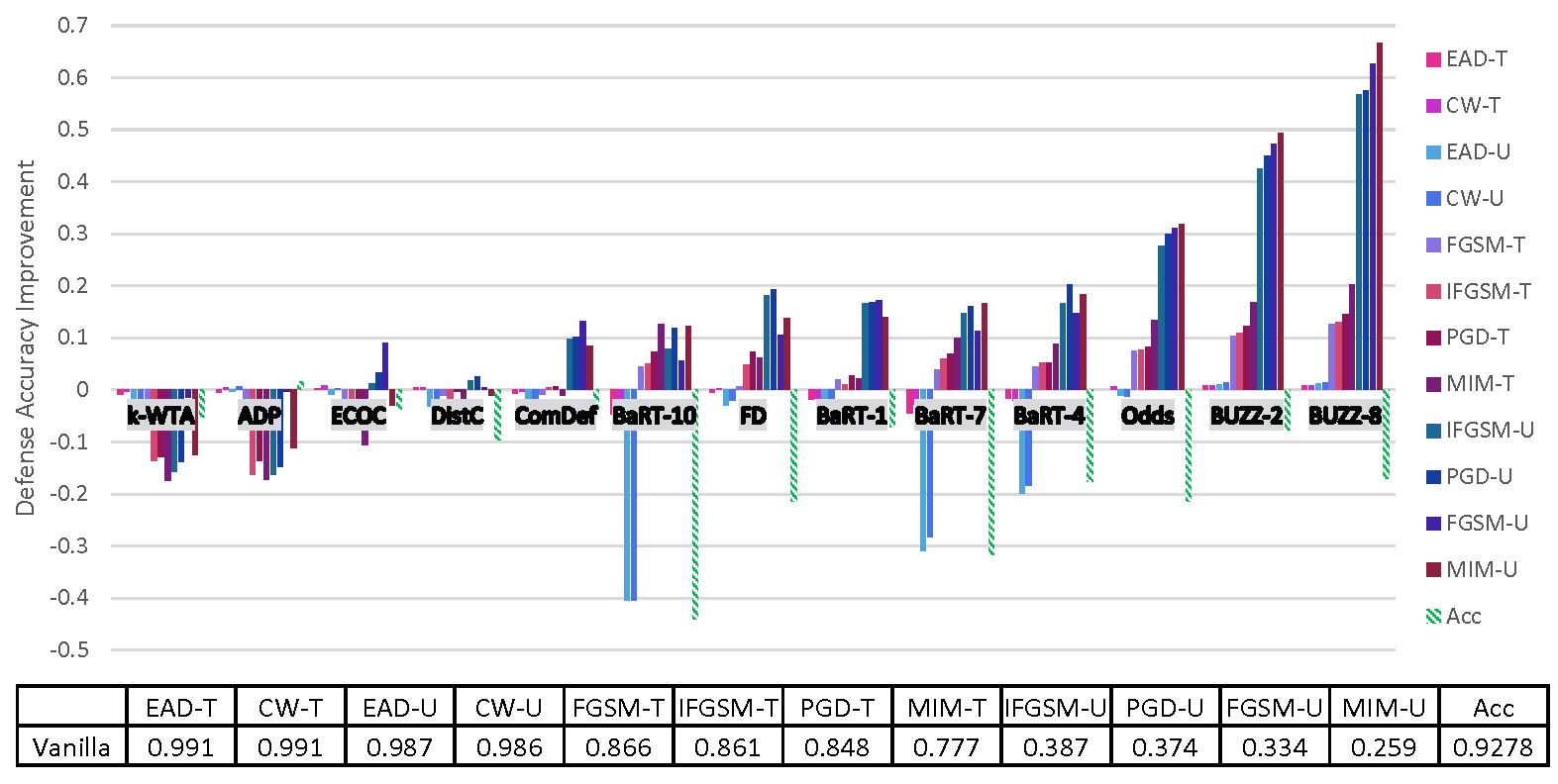}
\caption[]{CIFAR-10 adaptive black-box attack on each defense. Here the U/T refers to whether the attack is untargeted/targeted. Negative values means the defense performs worse than the no defense (vanilla) case. The Acc value refers to the drop in clean accuracy incurred by implementing the defense. The chart below the graph gives the vanilla defense accuracy numbers.}
 \label{fig:cifar10Mixed}
 \noindent
\end{figure*}

\begin{figure*}[!htb]
\centering
\includegraphics[scale=0.8]{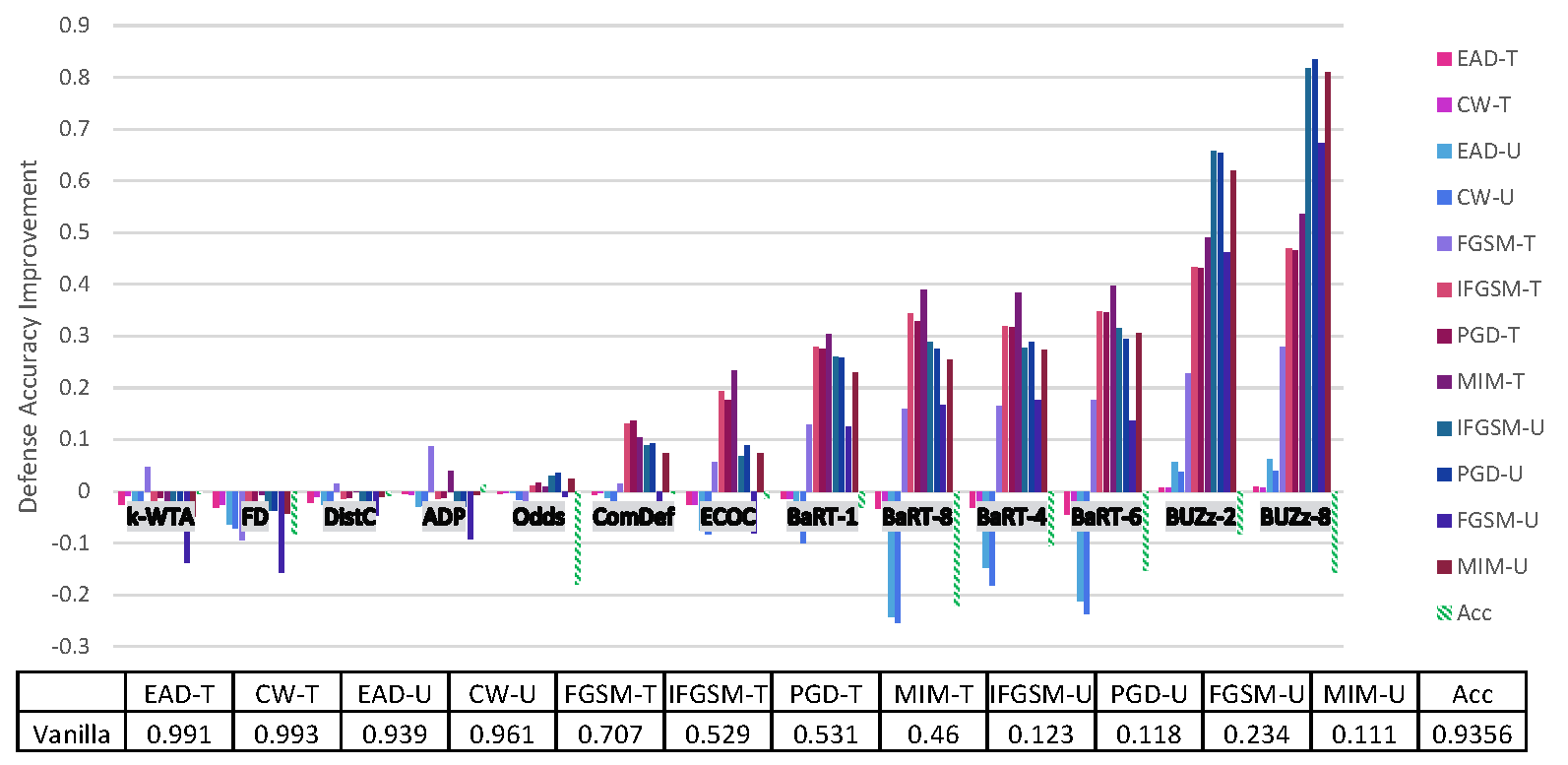}
\caption[]{Fashion-MNIST adaptive black-box attack on each defense. Here the U/T refers to whether the attack is untargeted/targeted. Negative values means the defense performs worse than the no defense (vanilla) case. The Acc value refers to the drop in clean accuracy incurred by implementing the defense. The chart below the graph gives the vanilla defense accuracy numbers.}
 \label{fig:fmnistMixed}
 \noindent
\end{figure*}

1. \textbf{Marginal or negligible improvements over no defense}: Figure~\ref{fig:cifar10Mixed} shows the defense results for CIFAR-10 with a $100\%$ strength adaptive black-box adversary. In this figure, we can clearly see 7 out of 9 defenses give marginal (less than $25\%$) increases in defense accuracy for any attack. BUZz and the Odds defense are the only ones to break this trend for CIFAR-10. For example, BUZz-8 gives a $66.7\%$ defense accuracy improvement for the untargeted MIM attack. Odds gives a $31.9\%$ defense accuracy improvement for the untargeted MIM attack. Likewise, for Fashion-MNIST again, 7 out of 9 defenses give only marginal improvements (see Figure~\ref{fig:fmnistMixed}). BUZz and BaRT are the exceptions for this dataset. 

2. \textbf{Security is not free (yet)}: Thus far, no defense we experimented with that offers significant (greater than $25\%$ increase) improvements comes for free. For example, consider the defenses that give significant defense accuracy improvements. BUZz-8 drops the clean accuracy by $17\%$ for CIFAR-10. BaRT-6 drops the clean accuracy by $15\%$ for Fashion-MNIST. As defenses improve, we expect to see this trade-off between clean accuracy and security become more favorable. However, our experiments show we have not reached this point with the current defenses.

3. \textbf{Common defense mechanisms}: It is difficult to decisively prove any one defense mechanism guarantees security. However, among the defenses that provide more than marginal improvements (Odds, BUZz and BaRT), we do see common defense trends. Both Odds and BUZz use adversarial detection. This indirectly deprives the adaptive black-box adversary of training data. When an input sample is marked as adversarial, the black-box attacker cannot use it to train the synthetic model. This is because the synthetic model has no adversarial class label. It is worth noting that in the appendix, we also argue why a synthetic model should not be trained to output an adversarial class label. 

Along similar lines, both BaRT and BUZz offer significant defense accuracy improvements for Fashion-MNIST. Both employ image transformations so jarring that the classifier must be retrained on transformed data. The experiments show that increasing the number of the transformations only increases security up to a certain point though. For example, BaRT-8 does not perform better than BaRT defenses that use less image transformations (see BaRT-6 and BaRT-4 in Figure~\ref{fig:fmnistMixed}).

\begin{figure*}[!htb]
\centering
\includegraphics[scale=0.8]{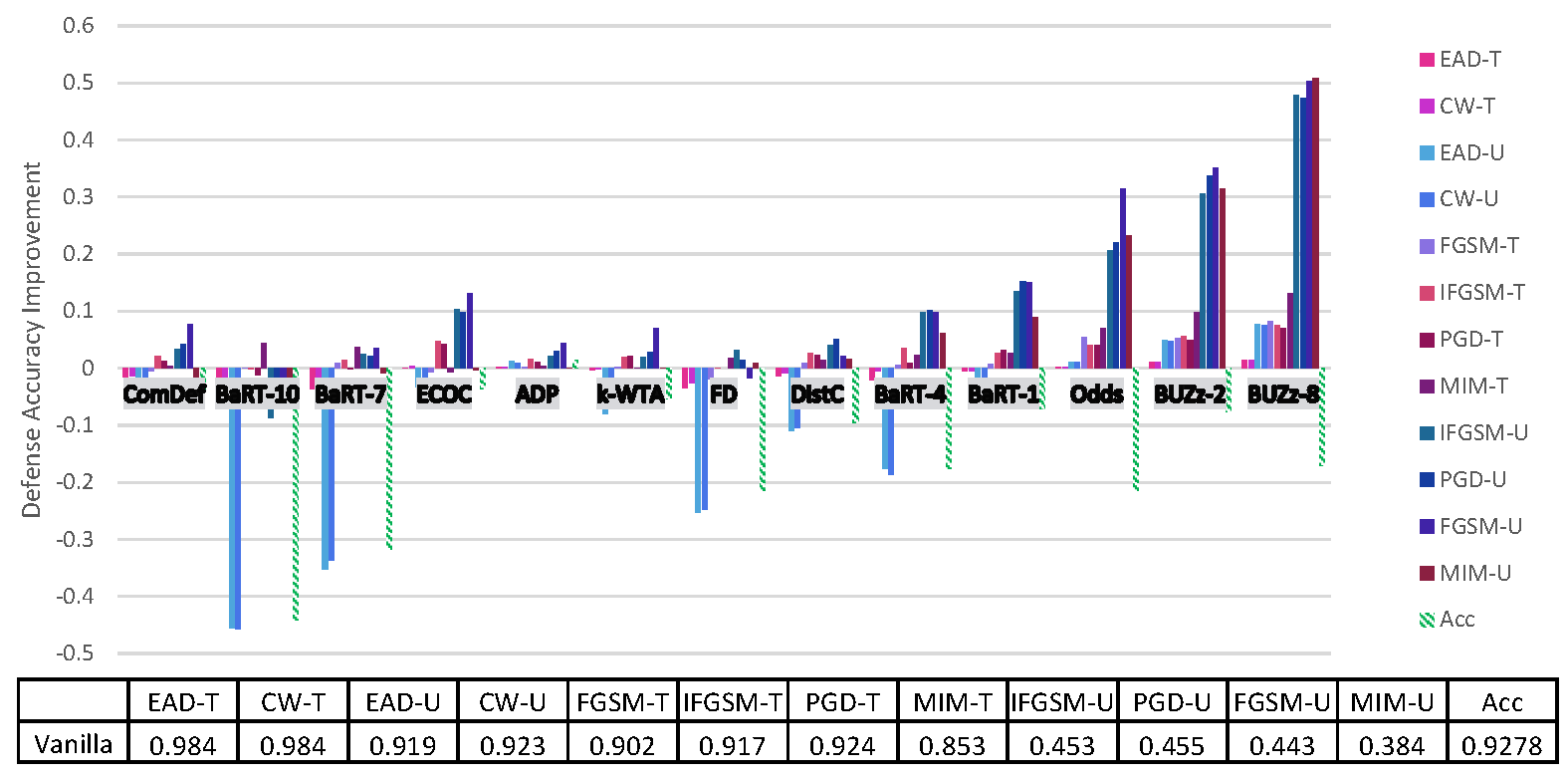}
\caption[]{CIFAR-10 pure black-box attack on each defense. Here the U/T refers to whether the attack is untargeted/targeted. Negative values means the defense performs worse than the no defense (vanilla) case. The Acc value refers to the drop in clean accuracy incurred by implementing the defense. The chart below the graph gives the vanilla defense accuracy numbers. For all the experimental numbers see Table~\ref{table:CIFARPure}.}
 \label{fig:pureCIFAR10}
\noindent
\end{figure*}

\begin{figure*}[!htb]
\centering
\includegraphics[scale=0.8]{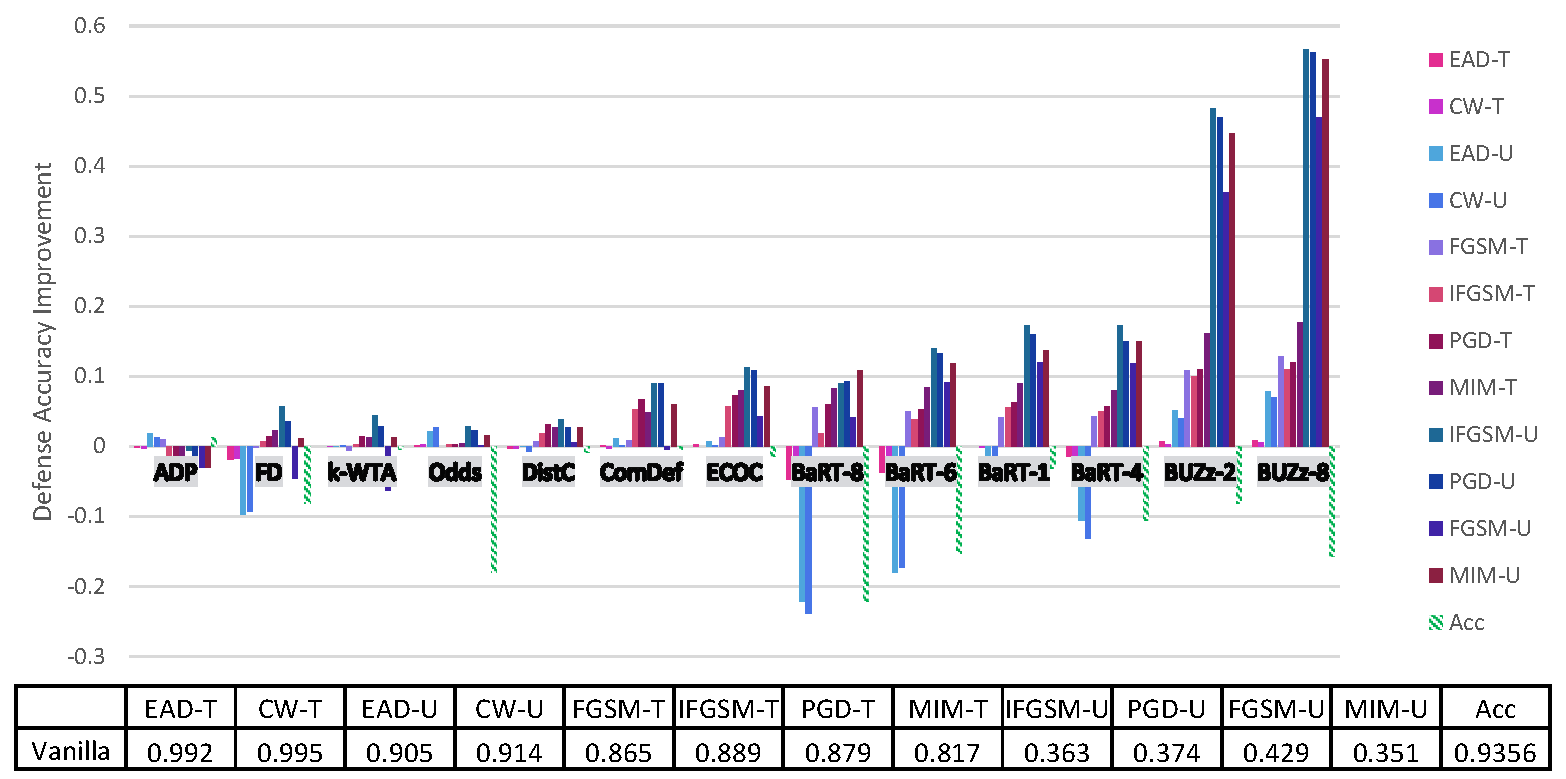}
\caption[]{Fashion-MNIST pure black-box attack on each defense. Here the U/T refers to whether the attack is untargeted/targeted. Negative values means the defense performs worse than the no defense (vanilla) case. The Acc value refers to the drop in clean accuracy incurred by implementing the defense. The chart below the graph gives the vanilla defense accuracy numbers. For all the experimental numbers see Table~\ref{table:FMPure}.}
 \label{fig:fMNISTPure}
\noindent
\end{figure*}

4. \textbf{Adaptive and pure black-box follow similar trends.} In Figure~\ref{fig:pureCIFAR10}, and ~\ref{fig:fMNISTPure} we show results for the pure black-box attack for CIFAR-10 and Fashion-MNIST. Just like for the adaptive black-box attack, we see similar trends in terms of which defenses provide the highest security gains. For CIFAR-10, the defenses that give at least $25\%$ greater defense accuracy than the vanilla defense include BUZz and Odds. For Fashion-MNIST, the only defense that gives this significant improvement is BUZz.

5. \textbf{Future defense analyses should be broad}: From our first point in this subsection, it is clear that a majority of these defenses give marginal improvements or less. This brings up an important question, what impact does our security study have for future defenses? The main lesson is future defense designers need to test against a broad spectrum of attacks. From the literature, we see the majority of the 9 defenses already considered white-box attacks like PGD or FGSM and some weak black-box attacks. However, in the face of adaptive attacks, these defenses perform poorly. Future defense analyses at the very least need white-box attacks \textit{and} adaptive black-box attacks. By providing our paper's results and code we hope to help future defense designers perform these analyses and advance the field of adversarial machine learning. 

\noindent

\noindent
\section{Individualized Experimental Defense Results}
\label{sec:indvResults}
In the previous section, we discussed the overarching themes represented in the adaptive black-box attack experimental results. In this section, we take a more fine grained approach and consider each defense individually.  

Both the $100\%$ adaptive black-box and pure black-box attack have access to the entire original training dataset. The difference between them lies in the fact that the adaptive black-box attack can generate synthetic data, and label the training data by querying the defense. Since both attacks are similar in terms of how much data they start with, a question arises. How effective is the attack if the attacker doesn't have access to the full training dataset? In the following subsections, we seek to answer that question by considering each defense under a variable strength adversary in the adaptive black-box setting. Specifically we test out adversaries that can query the defense but only have $75\%$, $50\%$, $25\%$ or $1\%$ of the original training dataset.

To simplify things with the variable strength adaptive black-box adversary, we only consider the untargeted MIM attack for generating adversarial examples. We use the MIM attack because it is the best performing attack on the vanilla (no defense) network for both datasets. Therefore, this attack represents the place where the most improvement in security can be made. For the sake of completeness, we do report all the defense accuracies for all six types of attacks for the variable strength adaptive black-box adversaries in the tables at the end of this section. 

\subsection{Barrage of random transforms analysis}

\begin{figure*}[!htb]
\centering
\includegraphics[scale=0.55]{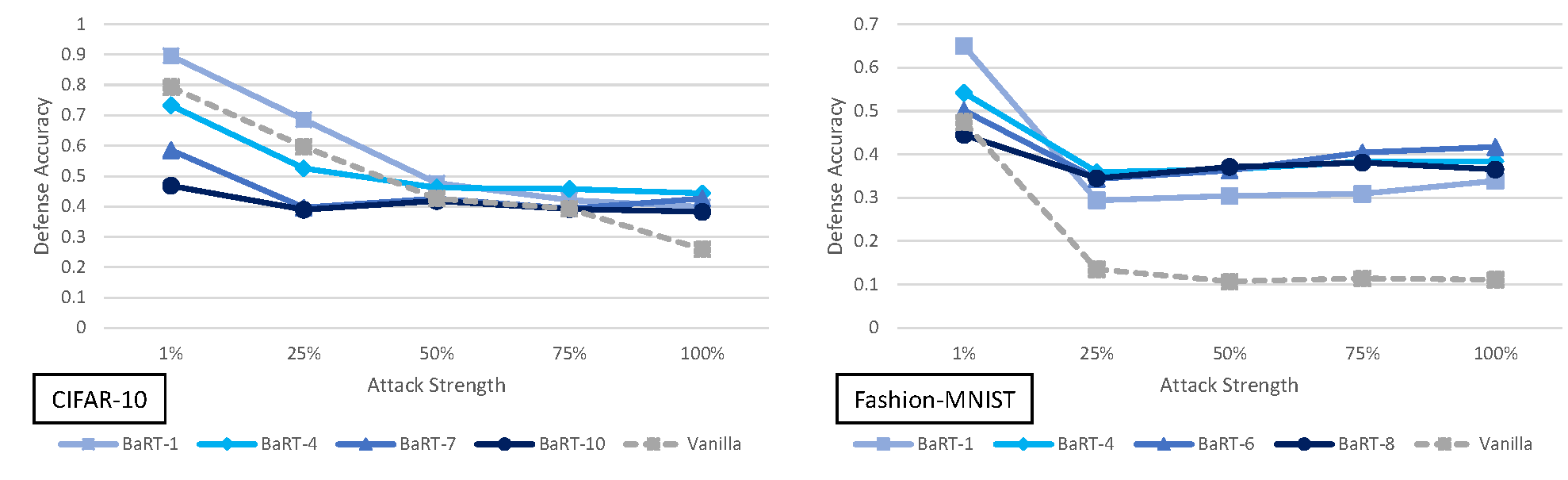}
\caption[]{Defense accuracy of barrage of random transforms defense on various strength adaptive black-box adversaries for CIFAR-10 and Fashion-MNIST. The defense accuracy in these graphs is measured on the adversarial samples generated from the untargeted MIM adaptive black-box attack. The $\%$ strength of the adversary corresponds to what percent of the original training dataset the adversary has access to. For full experimental numbers for CIFAR-10, see Tables 7 through 11. For full experimental numbers for Fashion-MNIST, see Tables 13 through 17.}
 \label{fig:bartadaptiveVar}
\noindent
\end{figure*}

The adaptive black-box attack with variable strength for BaRT defenses is shown in Figure~\ref{fig:bartadaptiveVar}. There are several interesting observations that can be made about this defense. First, for CIFAR-10, the maximum transformation defense (BaRT-10) actually performs worse than the vanilla defense in most cases. BaRT-1, BaRT-4 and BaRT-7 perform approximately the same as the vanilla defense. These statements hold except for the $100\%$ strength adaptive black-box adversary. Here, all BaRT defenses show a $12\%$ or greater improvement over the vanilla defense. 

Where as the performance of BaRT is rather varied for CIFAR-10, for Fashion-MNIST this is not the case. All BaRT defenses show improvement for the MIM attack for adversaries with $25\%$ strength or greater. 

When examining the results of BaRT on CIFAR-10 and Fashion-MNIST, we see a clear discrepancy in performance. One possible explanation is as follows: the image transformations in a defense must be selected in a way that does not greatly impact the original clean accuracy of the classifier. In the case of BaRT-10 (the maximum number of transformations) for CIFAR-10, it performs much worse than the vanilla case. However, BaRT-8 for Fashion-MNIST (again the maximum number of transformations) performs much better than the vanilla case. If we look at the clean accuracy of BaRT-10, it is approximately $48\%$ on CIFAR-10. This is a drop of more than $40\%$ as compared to the vanilla clean accuracy. For BaRT-8, the clean accuracy is approximately $72\%$ on Fashion-MNIST which is a drop of about $21\%$. Here we do not use precise numbers when describing the clean accuracy because as a randomized defense, the clean accuracy may drop or rise a few percentage points every time the test set is evaluated. 

From the above stated results, we can make the following conclusion: A defense that employs random image transformations cannot be applied naively to every dataset. The set of image transformations must be selected per dataset in such a manner that the clean accuracy is not drastically impacted. In this sense, while random image transformations may be a promising defense direction, it seems they may need to be designed on a per dataset basis. 

\noindent

\subsection{End-to-end image compression models analysis}

\begin{figure*}[!htb]
\centering
\includegraphics[scale=0.55]{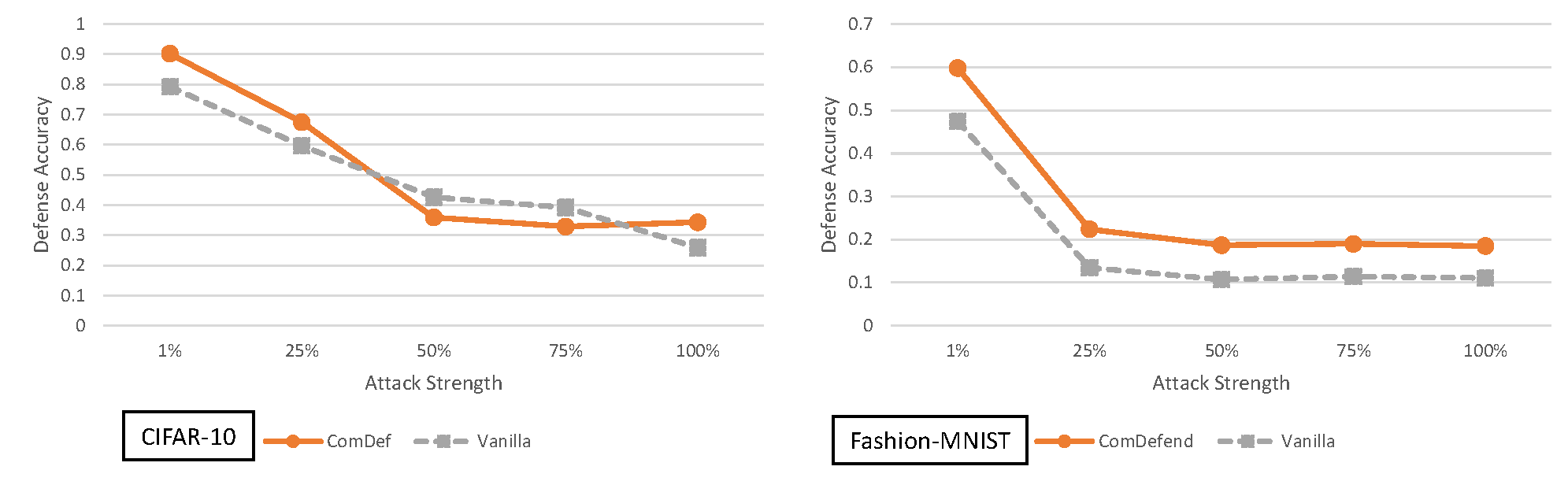}
\caption[]{Defense accuracy of ComDefend on various strength adaptive black-box adversaries for CIFAR-10 and Fashion-MNIST. The defense accuracy in these graphs is measured on the adversarial samples generated from the untargeted MIM adaptive black-box attack. The $\%$ strength of the adversary corresponds to what percent of the original training dataset the adversary has access to. For full experimental numbers for CIFAR-10, see Tables 7 through 11. For full experimental numbers for Fashion-MNIST, see Tables 13 through 17.}
 \label{fig:comdefVaradaptive}
\noindent
\end{figure*}

The adaptive black-box attack with variable strength for ComDefend is shown in Figure~\ref{fig:comdefVaradaptive}. For CIFAR-10, we see the defense performs very close to the vanilla network (and sometimes slightly worse). On the other hand, for Fashion-MNIST, the defense does offer a modest average defense accuracy improvement of $8.84\%$ across all adaptive black-box adversarial models.  

In terms of understanding the performance of ComDefend, it is important to note the following: In general it has been shown that more complex architectures (e.g., deeper networks) can better resist transfer based adversarial attacks~\cite{LiuChenLiuEtAl2016}. In essence an autoencoder/decoder setup can be viewed as additional layers in the CNN and hence a more complex model. Although this concept was shown for ImageNet~\cite{LiuChenLiuEtAl2016}, it may be a phenomena that occurs in other datasets as well. 

This more complex model can partially explain why ComDefend slightly outperforms the vanilla defense in most cases. In short, a slightly more complex model is slightly more difficult to learn and attack. Of course this begs the question, if a more complex model yields more security, why does the model complexity even have come from an autoencoder/decoder? Why not use ResNet164 or ResNet1001? 

These are all valid questions which are possible directions of future studies. While ComDefend itself does not yield significant (greater than $25\%$) improvements in security, it does bring up an interesting question: Under a black-box adversarial model, to what extent can increasing model complexity also increase defense accuracy? We leave this as an open question for possible future work. 

\noindent

\subsection{The odds are odd analysis}

\begin{figure*}[!htb]
\centering
\includegraphics[scale=0.55]{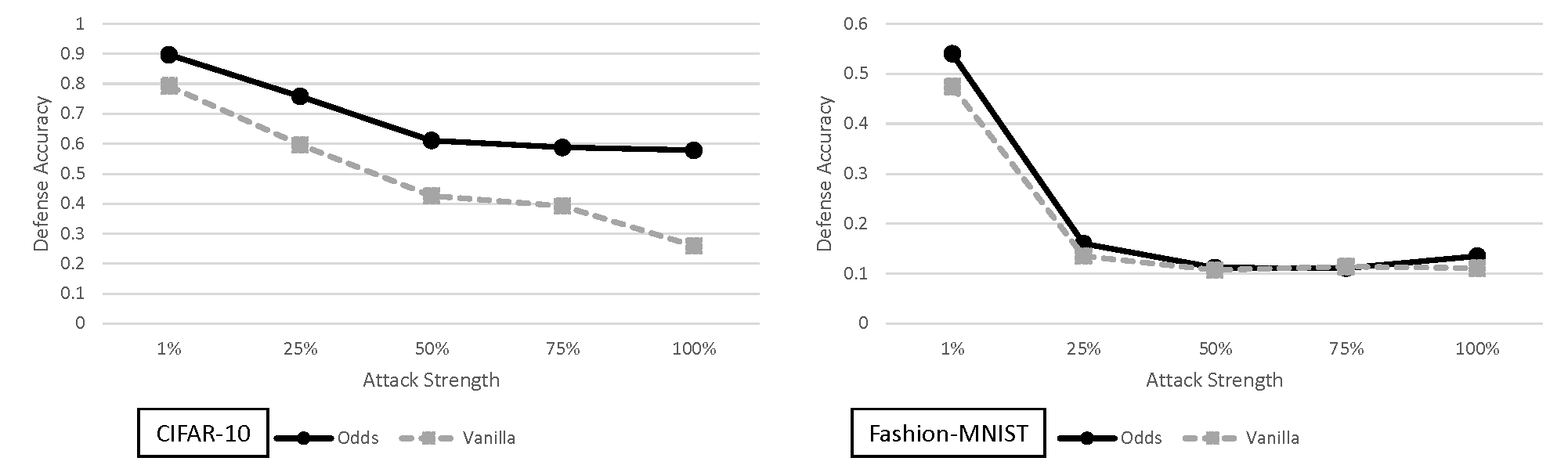}
\caption[]{Defense accuracy of the odds defense on various strength adaptive black-box adversaries for CIFAR-10 and Fashion-MNIST. The defense accuracy in these graphs is measured on the adversarial samples generated from the untargeted MIM adaptive black-box attack. The $\%$ strength of the adversary corresponds to what percent of the original training dataset the adversary has access to. For full experimental numbers for CIFAR-10, see Tables 7 through 11. For full experimental numbers for Fashion-MNIST, see Tables 13 through 17.}
 \label{fig:oddadaptiveVar}
\noindent
\end{figure*}

In Figure~\ref{fig:oddadaptiveVar}, the adaptive black-box attack with different strengths is shown for the Odds defense. For CIFAR-10 the Odds has an average improvement of $19.3\%$ across all adversarial models. However, for Fashion-MNIST the average improvement over the vanilla model is only $2.32\%$. As previously stated, this defense relies on the underlying assumption that creating a test for one set of adversarial examples will then generalize to all adversarial examples. 

When the test used in the Odds does provide security improvements (as in the case for CIFAR-10), it does highlight one important point. If the defense can mark some samples as adversarial, it is possible to deprive the adaptive black-box adversary of data to train the synthetic model. This in turn weakens the overall effectiveness of the adaptive black-box attack. We stress however that this occurs only when the test is accurate and does not greatly hurt the clean prediction accuracy of the classifier.    

\noindent

\subsection{Feature distillation analysis}

\begin{figure*}[!htb]
\centering
\includegraphics[scale=0.55]{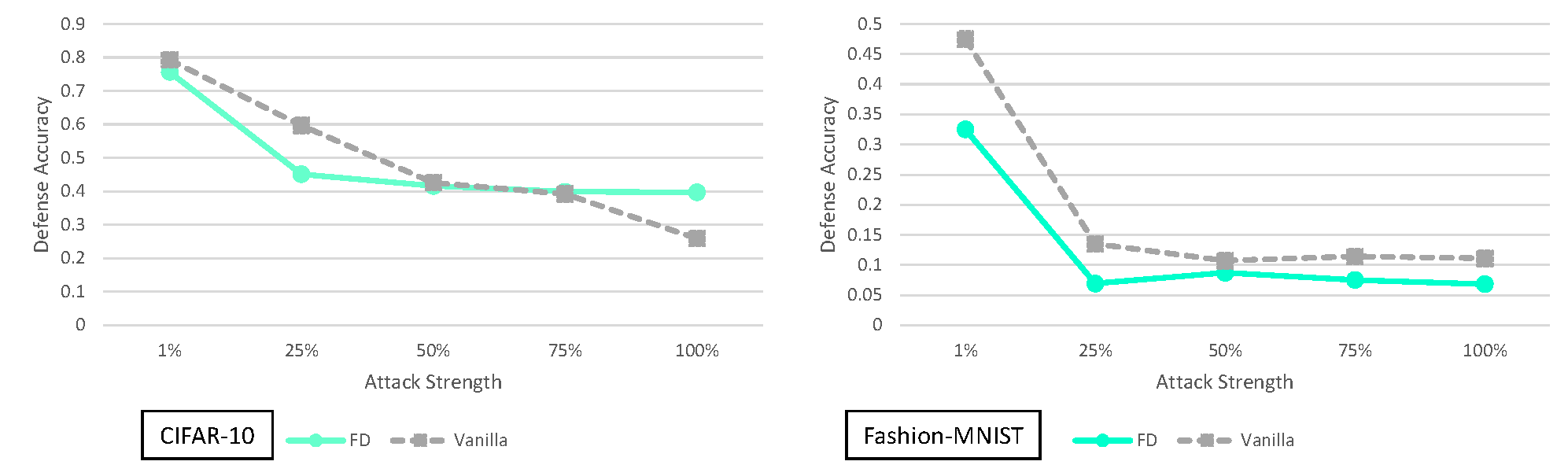}
\caption[]{Defense accuracy of feature distillation on various strength adaptive black-box adversaries for CIFAR-10 and Fashion-MNIST. The defense accuracy in these graphs is measured on the adversarial samples generated from the untargeted MIM adaptive black-box attack. The $\%$ strength of the adversary corresponds to what percent of the original training dataset the adversary has access to. For full experimental numbers for CIFAR-10, see Tables 7 through 11. For full experimental numbers for Fashion-MNIST, see Tables 13 through 17.}
 \label{figFeatureDistadaptiveVar}
\noindent
\end{figure*}

Figure~\ref{figFeatureDistadaptiveVar} shows the adaptive black-box with a variable strength adversary for the feature distillation defense. In general feature distillation performs worse than the vanilla network for all Fashion-MNIST adversaries. It performs worse or roughly the same for all CIFAR-10 adversaries, except for the $100\%$ case where it shows a marginal improvement of $13.8\%$.

In the original feature distillation paper the authors claim that they test a black-box attack. However, our understanding of their black-box attack experiment is that the synthetic model used in their experiment was not trained in an adaptive way. To be specific, the adversary they use does not have query access to the defense. Hence, this may explain why when an adaptive adversary is considered, the feature distillation defense performs roughly the same as the vanilla network. 

As we stated in the main paper, it seems unlikely a single image transformation would be capable of providing significant defense accuracy improvements. Thus far, the experiments on feature distillation support that claim for the JPEG compression/decompression transformation. The study of this image transformation and the defense are still very useful. The idea of JPEG compression/decompression when combined with other image transformations may still provide a viable defense, similar to what is done in BaRT. 

\noindent

\subsection{Buffer zones analysis}

\begin{figure*}[!htb]
\centering
\includegraphics[scale=0.5]{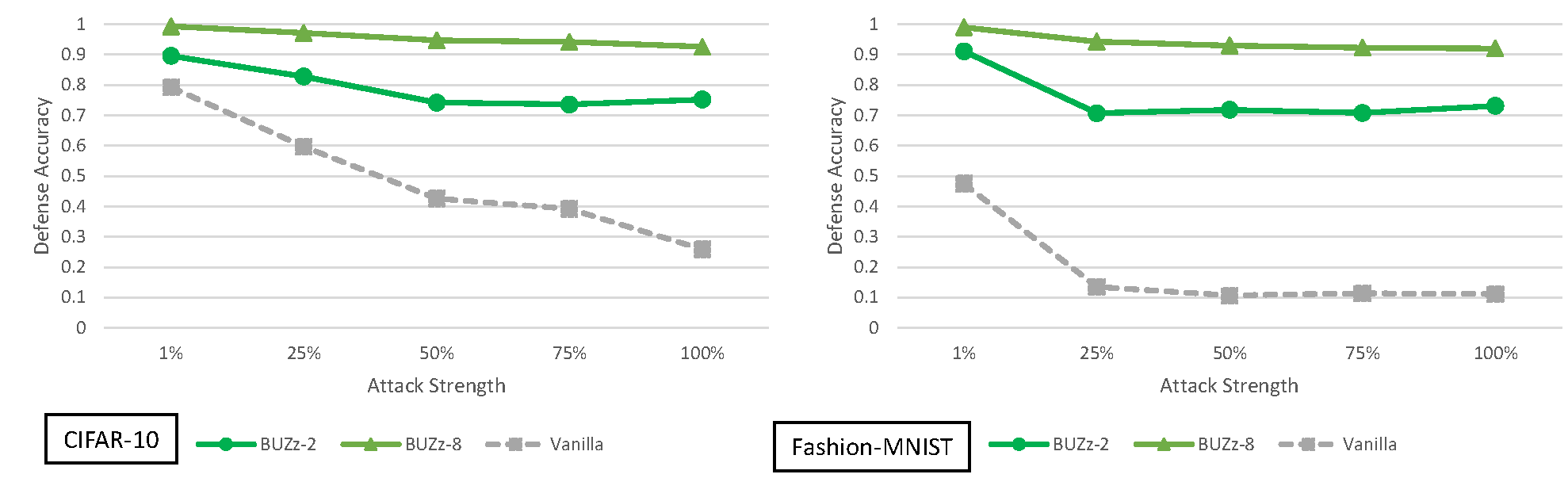}
\caption[]{Defense accuracy of the buffer zones defense on various strength adaptive black-box adversaries for CIFAR-10 and Fashion-MNIST. The defense accuracy in these graphs is measured on the adversarial samples generated from the untargeted MIM adaptive black-box attack. The $\%$ strength of the adversary corresponds to what percent of the original training dataset the adversary has access to. For full experimental numbers for CIFAR-10, see Tables 7 through 11. For full experimental numbers for Fashion-MNIST, see Tables 13 through 17.}
\noindent
\label{fig:bufferZoneadaptiveVar}
\end{figure*}

The results for the buffer zone defense in regards to the adaptive black-box variable strength adversary are given in Figure~\ref{fig:bufferZoneadaptiveVar}. For all adversaries, and all datasets we see an improvement over the vanilla model. This improvement is quite small for the $1\%$ adversary for the CIFAR-10 dataset at only a $10.3\%$ increase in defense accuracy for BUZz-2. However, the increases are quite large for stronger adversaries. For example, the difference between the BUZz-8 and vanilla model for the Fashion-MNIST full strength adversary is $80.9\%$.

As we stated earlier, BUZz is one of the defenses that does provide more than marginal improvements in defense accuracy. This improvement comes at a cost in clean accuracy however. To illustrate: BUZz-8 has a drop of $17.13\%$ and $15.77\%$ in clean testing accuracy for CIFAR-10 and Fashion-MNIST respectively. An ideal defense is one in which the clean accuracy is not greatly impacted. In this regard, BUZz still leaves much room for improvement. The overall idea presented in BUZz of combining adversarial detection and image transformations does give some indications of where future black-box security may lie, if these methods can be modified to better preserve clean accuracy.   

\noindent

\subsection{Improving adversarial robustness via promoting ensemble diversity analysis}

\begin{figure*}[!htb]
\centering
\includegraphics[scale=0.5]{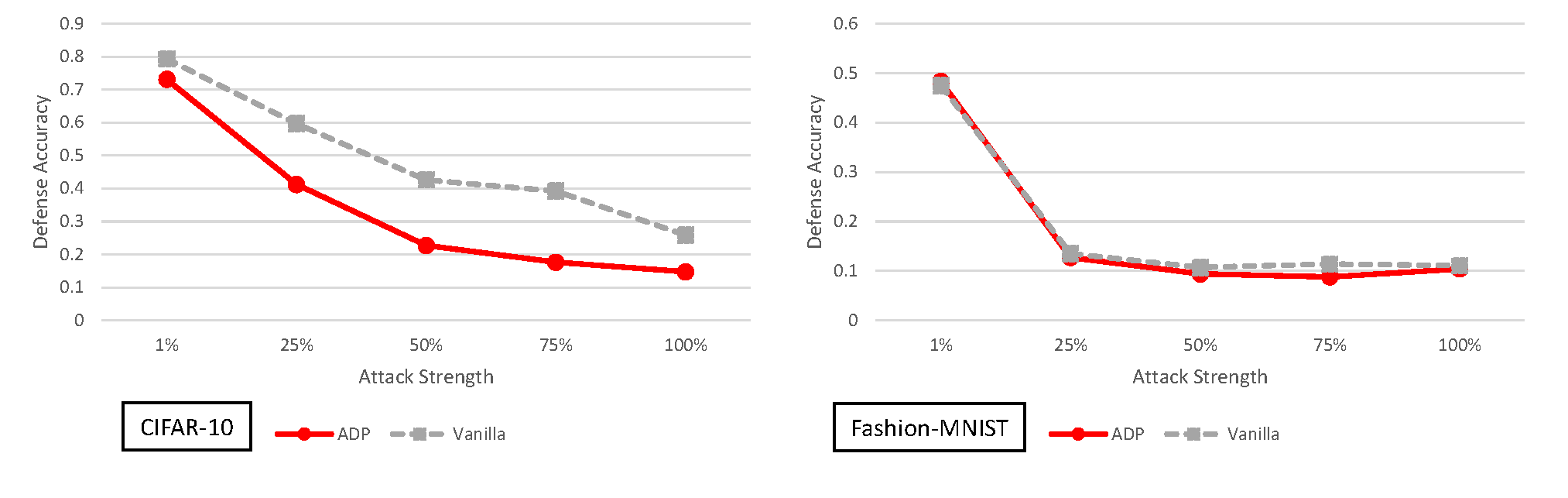}
\caption[]{Defense accuracy of the ensemble diversity defense on various strength adaptive black-box adversaries for CIFAR-10 and Fashion-MNIST. The defense accuracy in these graphs is measured on the adversarial samples generated from the untargeted MIM adaptive black-box attack. The $\%$ strength of the adversary corresponds to what percent of the original training dataset the adversary has access to. For full experimental numbers for CIFAR-10, see Tables 7 through 11. For full experimental numbers for Fashion-MNIST, see Tables 13 through 17.}
 \label{adpadaptiveVar}
\noindent
\end{figure*}
\noindent

The ADP defense and its performance under various strength adaptive black-box adversaries is shown in Figure~\ref{adpadaptiveVar}. For CIFAR-10, the defense does slightly worse than the vanilla model. For Fashion-MNIST, the defense does almost the same as the vanilla model. 

It has also been shown before in~\cite{nguyen2019buzz} that using multiple vanilla networks does not yield significant security improvements against a black-box adversary. The adaptive black-box attacks presented here support these claims when it comes to the ADP defense. At this time we do not have an adequate explanation as to why the ADP defense performs worse on CIFAR-10 given its clean accuracy is actually slightly higher than the vanilla model. We would expect slightly higher clean accuracy would result in slightly higher defense accuracy but this is not the case. Overall though, we do not see significant improvements in defense accuracy when implementing ADP against adaptive black-box adversaries of varying strengths for CIFAR-10 and Fashion-MNIST.  

\subsection{Enhancing transformation-based defenses against adversarial attacks with a distribution classifier analysis}

\begin{figure*}[!htb]
\centering
\includegraphics[scale=0.60]{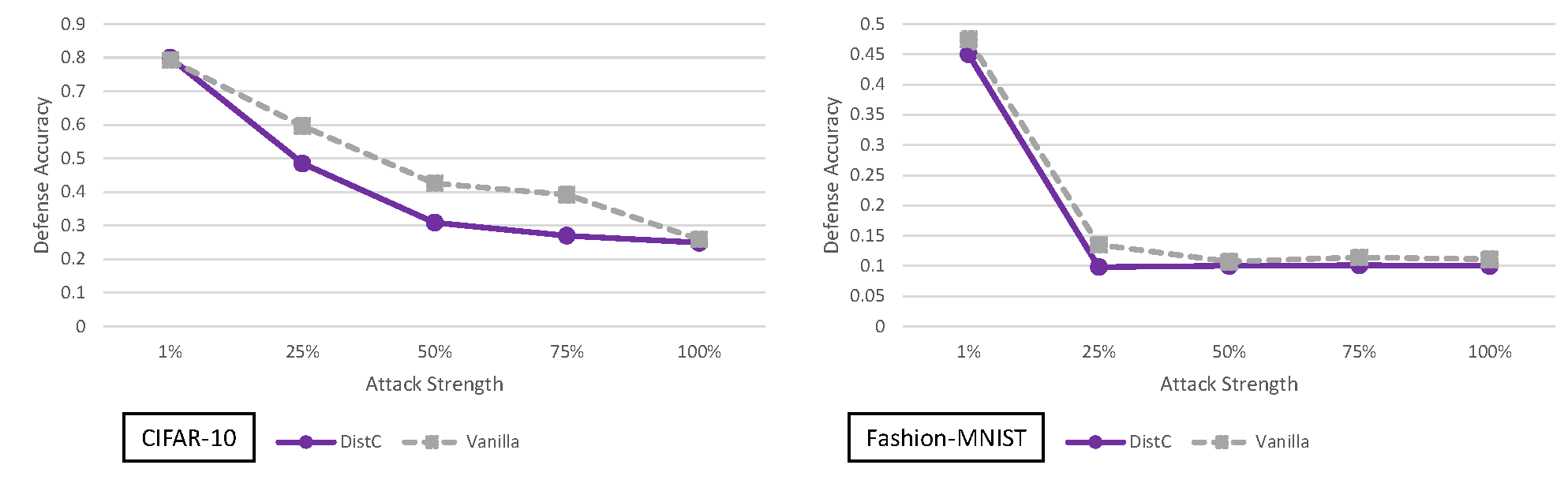}
\caption[]{Defense accuracy of the distribution classifier defense on various strength adaptive black-box adversaries for CIFAR-10 and Fashion-MNIST. The defense accuracy in these graphs is measured on the adversarial samples generated from the untargeted MIM adaptive black-box attack. The $\%$ strength of the adversary corresponds to what percent of the original training dataset the adversary has access to. For full experimental numbers for CIFAR-10, see Tables 7 through 11. For full experimental numbers for Fashion-MNIST, see Tables 13 through 17.}
 \label{fig:DistCadaptive}
\noindent
\end{figure*}

The distribution classifier defense~\cite{Kou2020Enhancing} results for adaptive black-box adversaries of varying strength are shown in Figure~\ref{fig:DistCadaptive}. This defense does not perform significantly better than the vanilla model for either CIFAR-10 or Fashion-MNIST. This defense employs randomized image transformations, just like BaRT. However, unlike BaRT, there is no clear improvement in defense accuracy. We can attribute this to two main reasons. First, the number of transformations in BaRT are significantly larger (i.e. 10 different image transformation groups in CIFAR-10, 8 different image transformation groups in Fashion-MNIST). In the distribution classifier defense, only resizing and zero padding transformations are used. Second, BaRT requires retraining the entire classifier to accommodate the transformations. This means all parts of the network from the convolutional layers, to the feed forward classifier are modified (retrained). The distribution classifier defense only retrains the final classifier after the soft-max output. This means the feature extraction layers (convolutional layers) between the vanilla model and the distributional classifier are virtually unchanged. If two networks have the same convolutional layers with the same weights, it is not surprising that the experiments show that they have similar defense accuracies.    

\noindent

\begin{figure*}[!htb]
\centering
\includegraphics[scale=0.55]{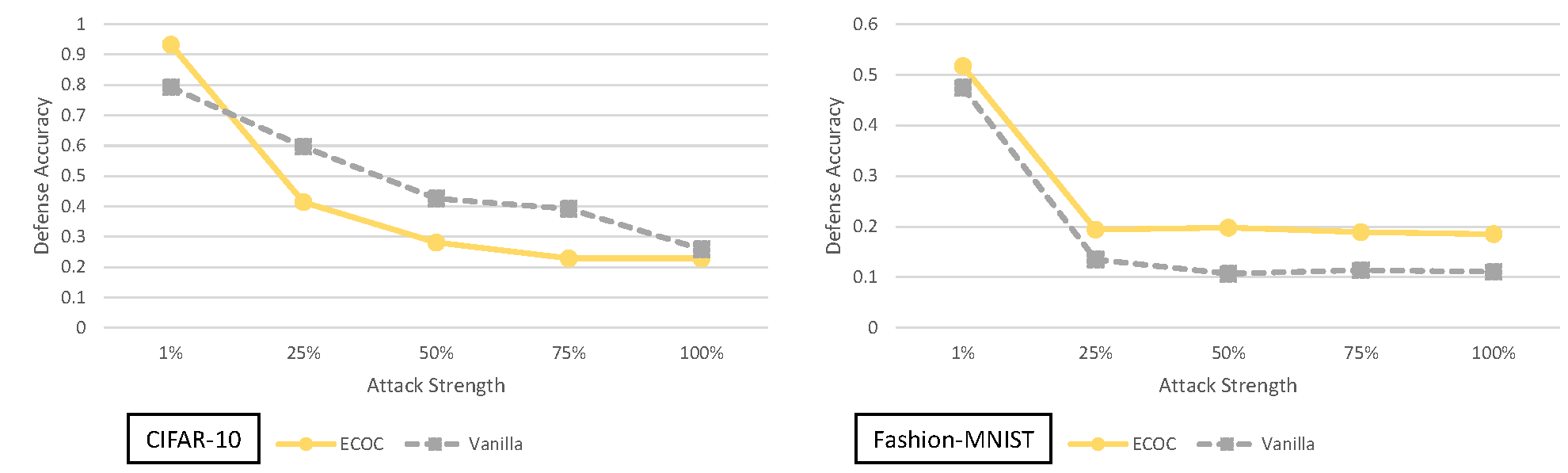}
\caption[]{Defense accuracy of the error correcting output code defense on various strength adaptive black-box adversaries for CIFAR-10 and Fashion-MNIST. The defense accuracy in these graphs is measured on the adversarial samples generated from the untargeted MIM adaptive black-box attack. The $\%$ strength of the adversary corresponds to what percent of the original training dataset the adversary has access to. For full experimental numbers for CIFAR-10, see Tables 7 through 11. For full experimental numbers for Fashion-MNIST, see Tables 13 through 17.}
 \label{fig:ECOCadaptiveVar}
\noindent
\end{figure*}
\noindent

\begin{figure*}[!htb]
\centering
\includegraphics[scale=0.55]{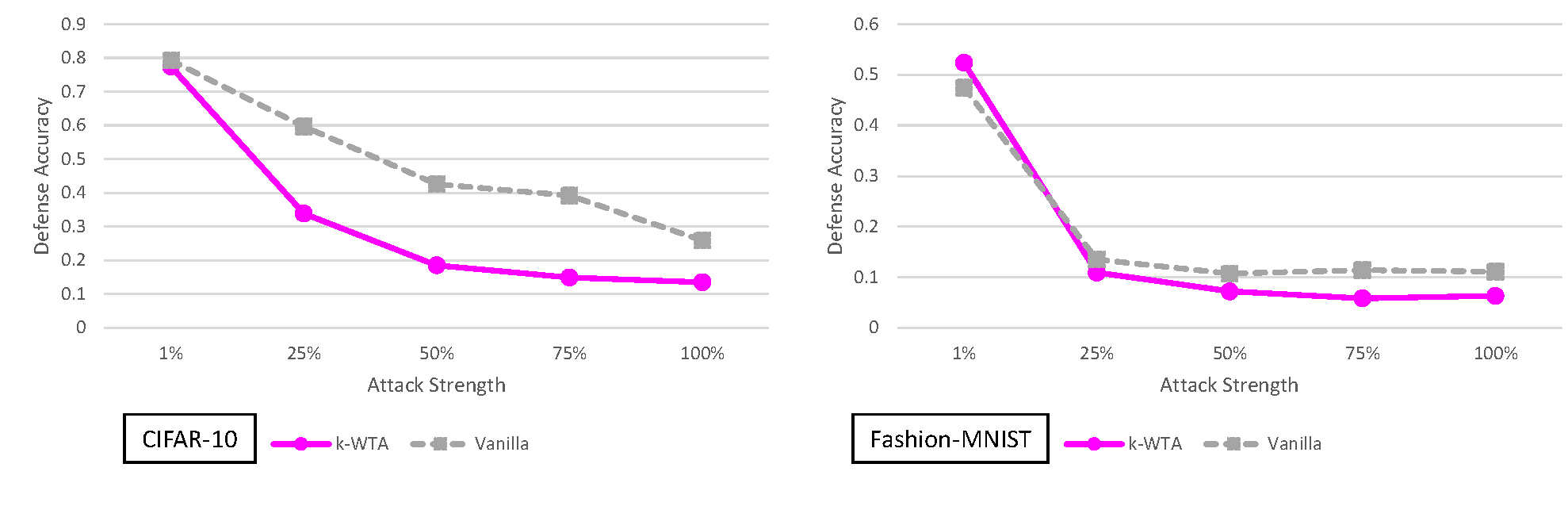}
\caption[]{Defense accuracy of the k-Winners-Take-All defense on various strength adaptive black-box adversaries for CIFAR-10 and Fashion-MNIST. The defense accuracy in these graphs is measured on the adversarial samples generated from the untargeted MIM adaptive black-box attack. The $\%$ strength of the adversary corresponds to what percent of the original training dataset the adversary has access to. For full experimental numbers for CIFAR-10, see Tables 7 through 11. For full experimental numbers for Fashion-MNIST, see Tables 13 through 17.}
 \label{fig:kWinneradaptiveVar}
\noindent
\end{figure*}
\noindent

\subsection{Error correcting output codes analysis}

In Figure~\ref{fig:ECOCadaptiveVar}, we show the ECOC defense for the adaptive black-box adversaries with varied strength. For CIFAR-10, ECOC performs worse than the vanilla defense in all cases except for the $1\%$ strength adversary. For Fashion-MNIST, the ECOC defense performs only slightly better than the vanilla model. ECOC performs $6.82\%$ greater in terms of defense accuracy on average when considering all the different strength adaptive black-box adversaries for Fashion-MNIST. In general, we don't see significant improvements (greater than $25\%$ increases) in defense accuracy when implementing ECOC.

\subsection{k-winner-take-all analysis}
The results for the adaptive black-box variable strength adversary for the k-WTA defense are given in Figure~\ref{fig:kWinneradaptiveVar}. We can see that the k-WTA defense performs approximately the same or slightly worse than the vanilla model in almost all cases. 

The slightly worse performance on CIFAR-10 can be attributed to the fact that the clean accuracy of the k-WTA ResNet56 is slightly lower than the clean accuracy of the vanilla model. We go into detailed explanations about the lower accuracy in the appendix. In short, the k-WTA defense is implemented in PyTorch while the vanilla ResNet56 is implemented in Keras. The slightly lower accuracy is due to implementation differences between Keras and PyTorch. It is not necessarily a direct product of the defense.

Regardless of the slight clean accuracy discrepancies, we see that this defense does not offer any significant improvements over the vanilla defense. From a black-box attacker perspective, this makes sense. Replacing an activation function in the network while still making it have almost identical performance on clean images should not yield security. The only exception to this would be if the architecture change fundamentally alters the way the image is processed in the CNN. In the case of k-WTA, the experiments support the hypothesis that this is not the case.

\section{Conclusion}
\label{sec:conclusion}
In this paper, we investigate and rigorously experiment with adaptive black-box attacks on recent defenses. Our paper's results span nine defenses, two adversarial models, six different attacks and two datasets. We show that most defenses (7 out of 9 for each dataset) offer less than a $25\%$ improvement in defense accuracy for an adaptive black-box adversary. Our analyses also yield new insight into what defense mechanisms are useful against black-box adversaries. Overall, we complete the security picture for currently proposed defense with our experiments and give future defense designers insight and direction with our analyses. 

\bibliographystyle{ACM-Reference-Format}
\bibliography{ref}
\clearpage
\appendix
\section{Appendix}
\subsection{Black-box settings}
We describe the detailed setup of our black box attacks in this paper.We strictly follow the setup of the black box attacks as described in~\cite{nguyen2019buzz}. This setup is carefully chosen by the authors to allow them to properly analyze the security of many defenses under the notion of pure black box attacks and adaptive black box attacks. For the sake of completeness, we re-introduce the setup used in~\cite{nguyen2019buzz}.

\begin{algorithm*}
  \caption{Construction of synthetic network $g$ in Papernot's oracle based black-box attack~\cite{nguyen2019buzz}}
  \label{alg:papernot_synthetic}
\begin{algorithmic}[1]
  \State {\textbf{Input}:} 
     \State \hspace{1cm} $\mathcal{O}$ represents black-box access to $F(f(\cdot))$ for target model $f$ with output function $F$;
     \State \hspace{1cm}  $\mathcal{X}_0\subseteq \mathcal{X}$, where $\mathcal{X}$ is the training data set of target model $f$;
     \State \hspace{1cm} substitute architecture $S$ 
     \State \hspace{1cm}  training method $\mathrm{M}$;
     \State \hspace{1cm}  constant $\lambda$;
     \State \hspace{1cm}  number $N$ of synthetic training epochs 
    \State {\bfseries Output:} 
      \State  \hspace{1cm}  synthetic model $s$ defined by parameters $\theta_{s}$ 
      \State \hspace{1cm} ($s$ also has output function $F$ which selects the max confidence score; 
      \State \hspace{1cm} $s$ fits architecture $S$)
      \State
      \State {\bfseries Algorithm:}
       \For{$N$ iterations}
          \State $\mathcal{D} \leftarrow \{(x,\mathcal{O}(x)): x \in \mathcal{X}_{t}\}$
          \State  $\theta_s = \mathrm{M}(S,\mathcal{D})$
          \State  $\mathcal{X}_{t+1} \leftarrow \{x+\lambda \cdot \mathrm{sgn}(J_{\theta_s}(x)[\mathcal{O}(x)]): x \in \mathcal{X}_t \} \cup \mathcal{X}_t$
       \EndFor
       \State Output $\theta_s$
\end{algorithmic}
\end{algorithm*}

Algorithm~\ref{alg:papernot_synthetic} describes the oracle based black-box attack from~\cite{PapernotMcDanielGoodfellowEtAl2017}. The oracle $\mathcal{O}$ represents black-box query access to the target model $f$ and only returns the final class label $F(f(x))$ for a query $x$ (and not the score vector $f(x)$).
Initially, the adversary is given (a part of) the training data set $\mathcal{X}$, i.e., he knows $\mathcal{D}=\{ (x, F(f(x))) : x\in \mathcal{X}_0\}$ for some $\mathcal{X}_0\subseteq \mathcal{X}$. 

Let $S$ and $\theta_s$ be an a-priori synthetic architecture and the parameter of the synthetic network, respectively. $\theta_s$ is trained using Algorithm~\ref{alg:papernot_synthetic}, i.e., the image-label pairs in $\mathcal{D}$ are used to train $\theta_s$ using a training method $M$ (e.g., Adam~\cite{kingma2014adam}). The data augmentation method (i.e., Jacobian) is used to increase the samples in training dataset $\mathcal{X}_t$ as described in line 17. Algorithm~\ref{alg:papernot_synthetic} runs $N$ iterations before outputting the final trained parameters $\theta_s$. 

\begin{table}[th!]
\caption{Training parameters used in the experiments~\cite{nguyen2019buzz}}
\label{tab:method}
\begin{center}
 \begin{tabular}{|c c|} 
 \hline
 Training Parameter & Value  \\ [0.5ex] 
 \hline\hline
 Optimization Method & ADAM  \\ 
 
 Learning Rate & 0.0001  \\
 
 Batch Size & 64  \\

 Epochs & 100  \\
 
 Data Augmentation & None  \\ [1ex] 
 \hline
\end{tabular}
\end{center}
\label{tab:exsetup}
\end{table}

\begin{table}[h!]
\caption{Adaptive black-box attack parameters~\cite{nguyen2019buzz}}
\centering
\scalebox{0.8}{
\begin{tabular}{|c|c|c|c|c|} 
 \hline
  & $|\mathcal{X}_0|$ & $N$ & $\lambda$  \\ [0.5ex] 
 \hline\hline
CIFAR-10& 50000 & 4 & 0.1 \\ \hline
Fashion-MNIST & 60000 & 4 & 0.1 \\ \hline 
\end{tabular}
}
\label{table:setup_datasets}
\end{table}

\begin{table}[th!]
\caption{Architectures of synthetic neural networks $s$ from \cite{CarliniWagner2016,nguyen2019buzz}}
\label{tab:CNN}
\begin{center}
 \begin{tabular}{|c c |} 
 \hline
 Layer Type & Fashion-MNIST and CIFAR-10 \\ [0.5ex] 
 \hline\hline
 Convolution + ReLU & 3 $\times$ 3 $\times$ 64 \\ 
 
 Convolution + ReLU & 3 $\times$ 3 $\times$ 64   \\
 
 Max Pooling & 2 $\times$ 2  \\
 
 Convolution + ReLU & 3 $\times$ 3 $\times$ 128  \\
 
 Convolution + ReLU & 3 $\times$ 3 $\times$ 128 \\

 Max Pooling & 2 $\times$ 2  \\
 
 Fully Connected + ReLU & 256  \\
 
 Fully Connected + ReLU & 256  \\
 
 Softmax & 10  \\ [1ex] 
 
 \hline
\end{tabular}
\end{center}
\label{tab:networksetup}
\end{table}

Tables \ref{tab:method}, \ref{table:setup_datasets}, and \ref{tab:CNN} from~\cite{nguyen2019buzz} describe the setup of our experiments in this paper. Table \ref{tab:method} presents the setup of the optimization algorithm used for training in Algorithm \ref{alg:papernot_synthetic}. The architecture of the synthetic model is described in Table~\ref{tab:CNN} and the main parameters for Algorithm~\ref{alg:papernot_synthetic} for CIFAR-10 and Fashion-MNIST are presented in Table \ref{table:setup_datasets}.

\subsection{The Adapative Black-Box Attack on null class label defenses}
For the adaptive black-box attack, there is a special case to consider when applying this attack to defenses that have the option of outputting a \textit{null class label}. We study two of these defenses, Odds and BUZz. Here we define the null class label $l$ as a label the defense gives to an input $x'$ when it considers the input to be manipulated by the adversary. This means for a 10 class problem like CIFAR-10, the defense actually has the option of outputting 11 class labels (with class label 11 being the adversarial label). In the context of the adaptive black-box attack, two changes occur. The first change is outside the control of the attacker, and is in regards to the definition of a successful attack. On a defense, that does not output a null class label, the attacker has to satisfy the following output condition: $\mathcal{O}(x')=y'$. We further specify $y'=y_{t}$ for a targeted attack or $y' \neq y$ for an untargeted attack. Also, we define $\mathcal{O}$ as the oracle in the defense, $x'$ as the adversarial example, $y$ as the original class label and $y_{t}$ as the target class label. The above formulation only holds when the defense does not employ any detection method (such as adversarial labeling). When adversarial labeling is employed the conditions change slightly. Now a successful attack must be misclassified by the defense and not be the null class label. Formally, we can write this as: $\mathcal{O}(x')=y' \land \mathcal{O}(x') \neq l$. While this first change is straightforward, there is another major change in the attack which we describe next.

The second main change in the adaptive black-box attack on null class label defenses comes from training the synthetic model. In the main paper, we mention that training the synthetic model is done with data labeled from the defense $\mathcal{O}(x)=y$. However, we do not use data which has a null class label $l$, i.e. $\mathcal{O}(x)=l$. We ignore this type of data because this would require modifying the untargeted attack in an unnecessary way. The untargeted attack  tries to find the malicious (wrong) label. If the synthetic network is outputting null labels, it is possible for the untargeted attack to produce an adversarial sample that will have a null label. In essence, the attack would fail under those circumstances. To prevent this, the objective function of every untargeted attack would need to be modified, such that the untargeted attack produces the malicious label and it is not the null label. To avoid needlessly complicating the attack, we simply do not use null labeled data. It is an open question of whether using null labeled data to train the synthetic network and the specialized untargeted attack we describe, would actually yield any meaningful performance gains.

\subsection{Vanilla model implementation}
\label{vanillaAppendix}
\textbf{CIFAR-10}: We train a ResNet56~\cite{He_2016} for 200 epochs with ADAM. We accomplish this using Keras\footnote{\url{https://github.com/keras-team/keras}} and the ResNet56 version 2 implementation\footnote{\url{https://keras.io/examples/cifar10_resnet/}}. In terms of the dataset, we use 50,000 samples for training and 10,000 samples for testing. All images are normalized in the range [0,1] with a shift of -0.5 so that they are in the range [-0.5, 0.5]. We also use the built in data augmentation technique provided by Keras during training. With this setup our vanilla network achieves a testing accuracy of $92.78\%$

\textbf{Fashion-MNIST}: We train a VGG16 network~\cite{simonyan2014very} for 100 epochs using ADAM. We use 60,000 samples for training and 10,000 samples for testing. All images are normalized in the range [0,1] with a shift of -0.5 so that they are in the range [-0.5, 0.5]. For this dataset we do not use any augmentation techniques. However, our VGG16 network has a built in resizing layer that transforms the images from 28x28 to 32x32. We found this process slightly boosts the clean accuracy of the network. On testing data we achieve an accuracy of $93.56\%$    

\subsection{Barrage of random transforms implementation}

The authors of BaRT~\cite{raff2019barrage} do not provide source code for their defense. We contacted the authors and followed their recommendations as closely as possible to re-implement their defense. However, some implementation changes had to be made. For the sake of the reproducibility of our results, we enumerate the changes made here. 

\textbf{Image transformations}: In the appendix for BaRT, they provide code snippets which are configured to work with scikit image package version 14.0.0.   However, due to compatibility issues, the closest version we could implement with our other existing packages was scikit image 14.4.0. Due to the different scikit version, two parts of the defense had to be modified. The original denoising wavelet transformation code in the BaRT appendix had invalid syntax for version 14.4.0, so we had to modify it and run it with different less random parameters.

The second defense change we made was due to error handling. In extremely rarely cases, certain sequences of image transformations return images with NAN values.  When contacting the authors they acknowledged that their code failed when using newer versions of sci-kit. As a result, in sci-kit 14.4.0 when we encounter this error, we randomly pick a new sequence of random transformations for the image. We experimentally verified that this has a negligible impact on the entropy of the defense. For example, in CIFAR-10 for the 5 transformation defense, we encounter this error 47 times when running all 50,000 training samples. That means roughly only 0.094$\%$ of the possible transformations sequences cannot be used in sci-kit 14.4.0. 

It is worth noting one other change we made to the Fashion-MNIST version of this defense. The original BaRT defense was only implemented for ImageNet, a three color channel (RGB) dataset. Fashion-MNIST is a single color channel (grayscale) dataset. As a results two transformation groups are not usable for the Fashion-MNIST BaRT defense (the color space change group and grayscale transformation group).

\textbf{Training BaRT:}
In~\cite{raff2019barrage} the authors start with a ResNet model pre-trained on ImageNet and further train it on transformed data for 50 epochs using ADAM. The transformed data is created by transforming samples in the training set. Each sample is transformed $T$ times, where $T$ is randomly chosen from distribution $U(0,5)$. Since the authors did not experiment with CIFAR-10 and Fashion-MNIST, we tried two approaches to maximize the accuracy of the BaRT defense. First, we followed the author's approach and started with a ResNet56 pre-trained for 200 epochs on CIFAR-10 with data-augmentation. We then further trained this model on transformed data for 50 epochs using ADAM. For CIFAR-10, we were able to achieve an accuracy of $98.87\%$ on the training dataset and a testing accuracy of $62.65\%$. Likewise, we tried the same approach for training the defense on the Fashion-MNIST dataset. We started with a VGG16 model that had already been trained with the standard Fashion-MNIST dataset for 100 epochs using ADAM. We then generated the transformed data and trained it for an additional 50 epochs using ADAM. We were able to achieve a $98.84\%$ training accuracy and a $77.80\%$ testing accuracy. Due to the relatively low testing accuracy on the two datasets, we tried a second way to train the defense.    

In our second approach we tried training the defense on the randomized data using untrained models. For CIFAR-10 we trained ResNet56 from scratch with the transformed data and data augmentation provided by Keras for 200 epochs. We found the second approach yielded a higher testing accuracy of $70.53\%$. Likewise for Fashion-MNIST, we trained a VGG16 network from scratch on the transformed data and obtained a testing accuracy of $80.41\%$. Due to the better performance on both datasets, we built the defense using models trained using the second approach.

\subsection{Improving adversarial robustness via promoting ensemble diversity implementation}

The original source code for the ADP defense~\cite{pang2019improving} on MNIST and CIFAR-10 datasets was provided on the author's Github page\footnote{\url{https://github.com/P2333/Adaptive-Diversity-Promoting}}. We used the same ADP training code the authors provided, but trained on our own architecture. For CIFAR-10, we used the ResNet56 model mentioned in subsection~\ref{vanillaAppendix} and for Fashion-MNIST, we used the VGG16 model mentioned in~\ref{vanillaAppendix}. We used K = 3 networks for ensemble model. We followed the original paper for the selection of the hyperparameters, which are $\alpha$ = 2 and $\beta$ = 0.5 for the adaptive diversity promoting (ADP) regularizer. In order to train the model for CIFAR-10, we trained using the 50,000 training images for 200 epochs with a batch size of 64. We trained the network using ADAM optimizer with Keras data augmentation. For Fashion-MNIST, we trained the model for 100 epochs with a batch size of 64 on the 60,000 training images. For this dataset, we again used ADAM as the optimizer but did not use any data augmentation. 

We constructed a wrapper for the ADP defense where the inputs are predicted by the ensemble model and the accuracy is evaluated. For CIFAR-10, we used 10,000 clean test images and obtained an accuracy of 94.3\%. We observed no drop in clean accuracy with the ensemble model, but rather observed a slight increase from $92.78\%$ which is the original accuracy of the vanilla model. For Fashion-MNIST, we tested the model with 10,000 clean test images and obtained an accuracy of $94.86\%$. Again for this dataset we observed no drop in accuracy after training with the ADP method.


\subsection{Error correcting output codes implementation} 

The training and testing code for ECOC defense~\cite{verma2019error} on CIFAR-10 and MNIST datasets was provided on the Github page of the authors\footnote{\url{https://github.com/Gunjan108/robust-ecoc/}}. We employed their “TanhEns32” method which uses 32 output codes and the hyperbolic tangent function as sigmoid function with an ensemble model. We choose this model because it yields better accuracy with clean and adversarial images for both CIFAR-10 and MNIST than the other ECOC models they tested, as reported in the original paper. 

For CIFAR-10, we used the original training code provided by the authors. Unlike the other defenses, we did not use a ResNet network for this defense because the models used in their ensemble predict individual bits of the error code. As a result these models are much less complex than ResNet56 (fewer trainable parameters). Due to the lower model complexity of each individual model in the ensemble, we used the default CNN structure the authors provided instead of our own. We did this to avoid over parameterization of the ensemble. We used 4 individual networks for the ensemble model and trained the network with 50,000 clean images for 400 epochs with a batch size of 200. We used data augmentation (with Keras) and batch normalization during training.


We used the original MNIST training code to train Fashion-MNIST by simply changing the dataset. Similarly, to avoid over parameterization, we again used the CNNs the authors used with lower complexity instead of using our VGG16 architecture. We trained the ensemble model with 4 networks for 150 epochs and with a batch size of 200. We did not use data augmentation for this dataset. 

For our implementation, we constructed our own wrapper class where the input images are predicted and evaluated using the TanhEns32 model. We tested the defense with 10,000 clean testing images for both CIFAR-10 and Fashion-MNIST, and obtained $89.08\%$ and $92.13\%$ accuracy, respectively.

\subsection{Distribution classifier implementation}


For the distribution classifier defense~\cite{Kou2020Enhancing}, we used random resize and pad (RRP)~\cite{XieWangZhangEtAl2017} and a DRN~\cite{DBLP:journals/corr/abs-1804-04775} as distribution classifier. The authors did not provide a public code  for their complete working defense. However, the DRN implementation by the same author was previously released on Github\footnote{\url{https://github.com/koukl/drn}}. We also contacted the authors, followed their recommendations for the training parameters and used the DRN implementation they sent to us as a blueprint.

In order to implement RRP, we followed the resize ranges the paper suggested, specifically for IFGSM  attack. Therefore, we chose the resize range as [19, 25] for CIFAR-10 and [22, 28] for Fashion-MNIST and used these parameters for all of our experiments. 

As for the distribution classifier, the DRN consists of fully connected layers and each node encodes a distribution. We use one hidden layer of 10 nodes. For the final layer, there are 10 nodes (representing each class) and there are two bins representing the logit output for each class. In this type of network the output from the layers are 2D. For the final classification, we convert from 2D to 1D by taking the output from the hidden layer and simply discarding the second bin each time. The distribution classifier then performs the final classification and outputs the class label.

\textbf{Training:} We followed the parameters the paper suggested to prepare training data. First, we collected 1000 correctly classified training clean images for Fashion-MNIST and 10,000 correctly classified clean images for CIFAR-10. Therefore, with no transformation, the accuracy of the networks is 100\%. For Fashion-MNIST, we used N = 100 transformation samples and for CIFAR-10, we used N = 50 samples, as suggested in the original paper. After collecting N samples from the RRP, we fed them into our main classifier network and collected the softmax probabilities for each class. Finally, for each class, we made an approximation by computing the marginal distributions using kernel density estimation with a Gaussian kernel (kernel width = 0.05). We used 100 discretization bins to discretize the distribution. For each image, we obtain 100 distribution samples per class. For further details of this distribution, we refer the readers to ~\cite{Kou2020Enhancing}.

We trained the model with the previously collected distribution of 1,000 correctly classified Fashion-MNIST images for 10 epochs as the authors suggested. For CIFAR-10, we trained the model with the distributions collected from 10,000 correctly classified images for 50 epochs. For both of the datasets, we used a learning rate of 0.1 and a batch size of 16. The cost function is the cross entropy loss on the logits and the distribution classifier is optimized using backpropagation with ADAM.\newline

\textbf{Testing:} We first tested the RRP defense alone with 10,000 clean test images for both CIFAR-10 and Fashion-MNIST to see the drop in clean accuracy. We observed that this defense resulted in approximately 71\% for CIFAR-10 and 82\% for Fashion-MNIST. Compared to the clean accuracies we obtain without the defense (93.56\% for Fashion-MNIST and 92.78\% for CIFAR-10), we observe drops in accuracy after random resizing and padding.

We tested the full implementation with RRP and DRN. In order to compare our results with the paper, we collected 5,000 correctly classified clean images for both datasets and collected distributions after transforming images using RRP (N = 50 for Fashion-MNIST and N = 100 for CIFAR-10) like we did for training. We observed a clean test accuracy of 87.48\% for CIFAR-10 and 97.76\% Fashion-MNIST, which is consistent with the results reported by the original paper. Clearly, if we test all of the clean testing data (10,000 images), we obtain lower accuracy (approximately 83\% for CIFAR-10 and 92\% for Fashion-MNIST) since there is also some drop in accuracy caused by the CNN. On the other hand, it can be seen that there is a smaller drop in clean accuracy as compared to the basic RRP implementation.

\subsection{Feature distillation implementation}

\textbf{Background}: The human visual system (HVS) is more sensitive to high frequency parts of the image and less sensitive to the low frequency parts. The standard JPEG compression is based on this understanding, so the standard JPEG quantization table compresses less sensitive frequency parts of the image (i.e. low frequency components) more than other parts. In order to defend against images, a higher compression rate is needed. However, since the CNNs work differently than the HVS, the testing accuracy and defense accuracy both suffer if a higher compression rate is used across all frequencies. In the Feature Distillation defense, as mentioned in Section \ref{sec:defsum}, a crafted quantization technique is used as a solution to this problem. A large quantization step ($QS$) can reduce adversarial perturbations but also cause more classification errors. Therefore, the proper selection of $QS$ is needed. In the crafted quantization technique, the frequency components are separated as Accuracy Sensitive (AS) band and Malicious Defense (MD) band. A higher quantization step ($QS_1$) is applied to the MD band to mitigate adversarial perturbations while a lower quantization step ($QS_2$) is used for AS band to enhance clean accuracy. For more details of this technique, we refer the readers to~\cite{liu2019feature}.

\textbf{Implementation}: The implementation of the defense can be found on the author's Github page\footnote{\url{https://github.com/zihaoliu123}}. However, this defense has only been implemented and tested for the ImageNet dataset by the authors. In order to fairly compare our results with the other defenses, we implemented and tested this defense for CIFAR-10 and Fashion-MNIST datasets.

This defenses uses two different methods: A one-pass process and a two-pass process. The one-pass process uses the proposed quantization/dequantization only in the decompression of the image. The two-pass process, on the other hand, uses the proposed quantization/dequantization in compression followed by one-pass process. In our experiments, we use the two-pass method as it has better defense accuracy than the one-pass process~\cite{liu2019feature}.

In the original paper, experiments were performed in order to find a proper selection of ($QS_1$) and ($QS_2$) for the AS and MD bands. At the end of these experiments, they set ($QS_1 = 30$) and ($QS_2 = 20$). However, these experiments were performed on ImageNet images where the images are much larger than CIFAR-10 and Fashion-MNIST images. Therefore, we performed experiments in order to properly select $QS_1$ and $QS_2$ for the Fashion-MNIST and CIFAR-10 datasets. For each dataset we start with the vanilla classifier (see~\ref{vanillaAppendix}). For each vanilla CNN we first do a one-pass and then generate 500 adversarial samples using untargeted FGSM. For CIFAR-10 we use $\epsilon=0.05$ and for Fashion-MNIST we use $\epsilon=0.15$. Here we use FGSM to do the hyperparameter selection for the defense because this is how the authors designed the original defense for ImageNet.

\begin{figure*}
\begin{adjustbox}{width=1\textwidth}
\subfloat[CIFAR-10]{\includegraphics[width = 2in]{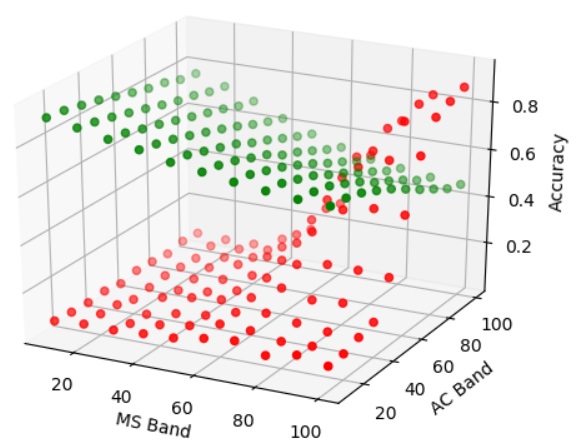}}
\subfloat[Fashion-MNIST]{\includegraphics[width = 2in]{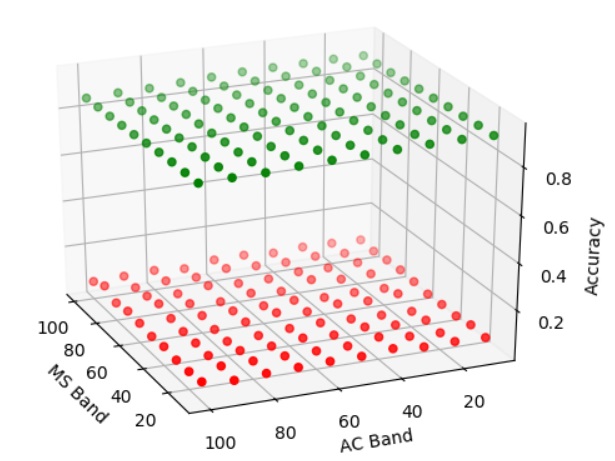}}
\end{adjustbox}
\caption{Feature distillation experiments to determine the hyperparameters for the defense. The x and y axis of the grid correspond to the specific hyperparameters for the defense. The Accuracy Sensitive band (denoted as AC in the figure) is the same as $QS_1$. The Malicious Defense band (denoted as MS in the figure) is the same as $QS_2$. On the z-axis the accuracy is measured. For every point in this grid two accuracy measurements are taken. The green dot corresponds to the clean accuracy using the QS values specified by the x-y coordinates. The red dot corresponds to the defense accuracy using the QS values specified by the x-y coordinates.}
\label{figureFD}
\end{figure*}

After generating the adversarial examples for each QS combination, we do a grid search over the possible hyperparameters $QS_1$ and $QS_2$. Specifically, we test 100 defense combinations by varying $QS_1$ from 10 to 100 and varying $QS_2$ from 10 to 100. For every possible combination of $QS_1$ and $QS_2$ we measure the accuracy on the clean test set and on the adversarial examples. The results of these experiments are shown in Figure~\ref{figureFD}.

In Figure~\ref{figureFD} for the CIFAR-10 dataset, there is an intersection where both the green dots and red dots overlap. This region represents a defense with both higher clean accuracy and higher defense accuracy (the idealized case). There are multiple different combinations of $QS_1$ and $QS_2$ that we could choose that give a decent trade-off. Here we arbitrarily select from among these better combinations $QS_1=70$ and $QS_2=40$ which gives a clean score of $71.4\%$ and a defense accuracy of $21.2\%$.

In Figure~\ref{figureFD} for the Fashion-MNIST dataset, there is no region in which both the clean accuracy and defense accuracy are high. This may show a limitation in the use of feature distillation as a defense for some datasets, as here no ideal trade-off exists. We pick $QS_1=70$ and $QS_2=40$ which gives a clean score of $89.34\%$ and a defense accuracy of $9\%$. We picked these values because this combination gave the highest defense accuracy out of all possible hyperparameter choices.

\subsection{End-to-end image compression models implementation}


The original source code for defenses on Fashion-MNIST and ImageNet  were provided by the authors of ComDefend\cite{jia2019comdefend} on their Github page\footnote{\url{https://github.com/jiaxiaojunQAQ/Comdefend}}. In addition, they included their trained compression and reconstruction models for Fashion-MNIST and CIFAR-10 separately.

Since this defense is a pre-processing module, it does not require modifications to the classifier network~\cite{jia2019comdefend}. Therefore, in order to perform the classification, we used our own models as described in Section~\ref{vanillaAppendix} and we combined them with this pre-processing module.

According to the authors of ComDefend, ComCNN and RecCNN were trained on 50,000 clean (not perturbed) images from the CIFAR-10 dataset for 30 epochs using a batch size of 50. In order to use their pre-trained models, we had to install the canton package v0.1.22 for Python. However, we had incompatibility issues with canton and the other Python packages installed in our system. Therefore, instead of installing this package directly, we downloaded the source code of the canton library from its Github page and added it to our defense code separately. We constructed a wrapper for ComDefend, where the type of dataset (Fashion-MNIST or CIFAR-10) is indicated as input so that the corresponding classifier can be used (either ResNet56 or VGG16). We tested the defense with the testin data of CIFAR-10 and Fashion-MNIST  and we were able to achieve an accuracy of $88\%$ and $93\%$ respectively.


\subsection{The odds are odd implementation}

\textbf{Mathematical background}: Here we give a detailed description of the defense based on the statistical test derived from the logits layer. For given image $x$, we denote $\phi(x)$ as the logits layer (i.e., the input to the softmax layer) of a classifier, $f_y= \langle w_y, \phi(x) \rangle$ where $w_y$ is the weight vector for the class $y, y \in \{1,\cdots,K\}$. The class label is determined by $F(x)= \text{argmax}_y f_y(x)$. We define pair-wise log-odds between class $y$ and $z$ as 

\begin{equation}\label{eq:log_odd}
    f_{y,z}(x) = f_z(x) - f_y(x) = \langle w_z - w_y, \phi(x) \rangle. 
\end{equation}  

We denote $f_{y,z}(x+\eta)$ the noise-perturbed log-odds where the noise $\eta$ is sampled from a distribution $\mathcal{D}$. Moreover, we define the following formulas for a pair $(y,z)$:

\begin{eqnarray}
g_{y,z} &:=& f_{y,z}(x+\eta) - f_{y,z}(x) \\ \nonumber
\mu_{y,z} &:=& \mathrm{E}_{x|y} \mathrm{E}_{\eta}[g_{y,z}(x,\eta)] \\ \nonumber
\sigma_{y,z}&:=&\mathrm{E}_{x|y}\mathrm{E}_{\eta}[(g_{y,z}(x,\eta)-\mu_{y,z})^2]\\ \nonumber
\bar{g}_{y,z}(x,\eta) &:=& [g_{y,z}(x,\eta)-\mu_{y,z}]/\sigma_{y,z}
\end{eqnarray}

For the original training data set, we compute $\mu_{y,z}$ and $\sigma_{y,z}$ for all $(y,z)$. We apply the untargeted white-box attack (PGD~\cite{madry2018towards}) to generate the adversarial dataset. After that, we compute $\mu_{y,z}^{adv}$ and $\sigma_{y,z}^{adv}$ using the adversarial dataset. We denote $\tau_{y,z}$ as the threshold to control the false positive rate (FPR) and it is computed based on $\mu_{y,z}^{adv}$ and $\sigma_{y,z}^{adv}$. The distribution of clean data and the distribution of adversarial data are represented by $(\mu,\sigma)$ and $(\mu^{adv},\sigma^{adv})$, respectively. These distributions are supposed to be separated and $\tau$ is used to control the FPR. 

For a given image $x$, the statistical test is done as follows. First, we calculate the expected perturbed log-odds $\bar{g}_{y,z}(x) = \mathrm{E}_{\eta}[\bar{g}_{y,z}(x,\eta)]$ where $y$ is the predicted class label of image $x$ given by the vanilla classifier. The test will determine the image $x$ with the label $y=F(x)$ as adversarial (malicious) if
\begin{equation*}
    \max_{z \neq y} \{\bar{g}_{y,z}(x)  - \tau_{y,z}\} \geq 0. 
\end{equation*}
Otherwise, the input will be considered benign. In case the test recognizes the image as malicious one, the ``corrected'' class label $z$ is defined as 

$$\max_{z}\{ \bar{g}_{y,z}(x)  - \tau_{y,z} \}.$$  

\textbf{Implementation details}: The original source code for the Odds defense~\cite{roth2019odds} on CIFAR-10 and ImageNet was provided by the authors\footnote{\url{https://github.com/yk/icml19_public}}. We use their code as a guideline for our own defense implementation. We develop the defense for the CIFAR-10 and  Fashion-MNIST and datasets. For each dataset, we apply the untargeted 10-iteration PGD attack on the vanilla classifier that will be used in the defense. Note this is a white-box attack. The parameters for the PGD attack are $\epsilon=0.005$ for CIFAR-10 and $\epsilon=0.015$ for Fashion-MNIST respectively. By applying the white-box PGD attack we can create the adversarial datasets for the defense. We choose these attack parameters because they yield adversarial examples with small noise. In~\cite{roth2019odds}, the authors assume that the adversarial examples are created by adding small noise. Hence, they are not robust against adding the white noises. For a given image, it is normalized first to be in the range $[-0.5,0.5]$. For each pixel, we generate a noise from $\mathcal{N}(0,0.05)$ and add it to the pixel.

For CIFAR-10, we create 50,000 adversarial examples. For Fashion-MNIST, we create 60,000 adversarial examples. We calculate $\mu,\sigma$ and $\tau$ for each data set for FPR=$1\%, 10\%,20\%,30\%,40\%,50\%$ and $80\%$ as described in the mathematical background. For each image, we evaluate it 256 times to compute $\bar{g}_{y,z}(x)$. Table~\ref{tab:clean_accuracy_odds_odd_cifar_fashionmnist} shows the prediction accuracy of the defense for the clean (non-adversarial) dataset for CIFAR-10 and Fashion-MNIST. To compute the clean prediction accuracy, we use 1000 samples from the test dataset of CIFAR-10 and Fashion-MNIST.

\begin{table*}[b!]
\caption{CIFAR-10 pure black-box attack. Note the defense numbers in the table are the defense accuracy minus the vanilla defense accuracy. This means they are relative accuracies. The very last row is the actual defense accuracy of the vanilla network.}
\centering
\label{table:CIFARPure}
\begin{adjustbox}{width=1\textwidth}
\begin{tabular}{|c|c|c|c|c|c|c|c|c|c|c|c|c|c|}
\hline
{\color[HTML]{000000} }                 & {\color[HTML]{000000} \textbf{FGSM-T}} & {\color[HTML]{000000} \textbf{IFGSM-T}} & {\color[HTML]{000000} \textbf{MIM-T}} & {\color[HTML]{000000} \textbf{PGD-T}} & {\color[HTML]{000000} \textbf{CW-T}} & {\color[HTML]{000000} \textbf{EAD-T}} & {\color[HTML]{000000} \textbf{FGSM-U}} & {\color[HTML]{000000} \textbf{IFGSM-U}} & {\color[HTML]{000000} \textbf{MIM-U}} & {\color[HTML]{000000} \textbf{PGD-U}} & {\color[HTML]{000000} \textbf{CW-U}} & {\color[HTML]{000000} \textbf{EAD-U}} & {\color[HTML]{000000} \textbf{Acc}} \\ \hline
{\color[HTML]{000000} \textbf{ADP}}     & {\color[HTML]{000000} 0.003}           & {\color[HTML]{000000} 0.016}            & {\color[HTML]{000000} 0.004}          & {\color[HTML]{000000} 0.011}          & {\color[HTML]{000000} 0.003}         & {\color[HTML]{000000} 0.003}          & {\color[HTML]{000000} 0.044}           & {\color[HTML]{000000} 0.022}            & {\color[HTML]{000000} 0.001}          & {\color[HTML]{000000} 0.03}           & {\color[HTML]{000000} 0.009}         & {\color[HTML]{000000} 0.013}          & {\color[HTML]{000000} 0.0152}       \\ \hline
{\color[HTML]{000000} \textbf{BaRT-1}}  & {\color[HTML]{000000} 0.007}           & {\color[HTML]{000000} 0.026}            & {\color[HTML]{000000} 0.027}          & {\color[HTML]{000000} 0.032}          & {\color[HTML]{000000} -0.005}        & {\color[HTML]{000000} -0.005}         & {\color[HTML]{000000} 0.151}           & {\color[HTML]{000000} 0.135}            & {\color[HTML]{000000} 0.089}          & {\color[HTML]{000000} 0.153}          & {\color[HTML]{000000} -0.07}         & {\color[HTML]{000000} -0.066}         & {\color[HTML]{000000} -0.0707}      \\
\hline
\end{tabular}
\end{adjustbox}
\end{table*}

\begin{table*}
\caption{CIFAR-10 adaptive black-box attack 1\%. Note the defense numbers in the table are the defense accuracy minus the vanilla defense accuracy. This means they are relative accuracies. The very last row is the actual defense accuracy of the vanilla network.}
\centering
\label{table:CIFARM1}
\begin{adjustbox}{width=1\textwidth}
\begin{tabular}{|c|c|c|c|c|c|c|c|c|c|c|c|c|c|}
\hline
                                        & \textbf{FGSM-T}              & \textbf{IFGSM-T}             & \textbf{MIM-T}               & \textbf{PGD-T}               & \textbf{FGSM-U}              & \textbf{IFGSM-U}             & \textbf{MIM-U}               & \textbf{PGD-U}               & \textbf{CW-T}                & \textbf{CW-U}                & \textbf{EAD-T}               & \textbf{EAD-U}               & \textbf{Acc}                  \\ \hline
\textbf{ADP}                            & -0.015                       & -0.001                       & -0.017                       & -0.013                       & -0.055                       & 0.002                        & -0.062                       & -0.007                       & 0.008                        & -0.001                       & 0.002                        & -0.003                       & 0.0152                        \\ \hline
\textbf{BaRT-1}                         & 0.011                        & 0.017                        & 0.013                        & 0.013                        & 0.125                        & 0.128                        & 0.103                        & 0.121                        & 0.009                        & -0.061                       & 0.009                        & -0.061                       & -0.0695                       \\ \hline
\textbf{BaRT-10}                        & -0.05                        & -0.044                       & -0.071                       & -0.042                       & -0.287                       & -0.277                       & -0.325                       & -0.28                        & -0.058                       & -0.439                       & -0.053                       & -0.394                       & -0.4408                       \\
\hline
\end{tabular}
\end{adjustbox}
\end{table*}

\begin{table*}
\caption{CIFAR-10 adaptive black-box attack 25\%. Note the defense numbers in the table are the defense accuracy minus the vanilla defense accuracy. This means they are relative accuracies. The very last row is the actual defense accuracy of the vanilla network.}
\centering
\label{table:CIFARM25}
\begin{adjustbox}{width=1\textwidth}
\begin{tabular}{|c|c|c|c|c|c|c|c|c|c|c|c|c|c|}
\hline
                                        & \textbf{FGSM-T}              & \textbf{IFGSM-T}             & \textbf{MIM-T}              & \textbf{PGD-T}               & \textbf{FGSM-U}              & \textbf{IFGSM-U}             & \textbf{MIM-U}               & \textbf{PGD-U}               & \textbf{CW-T}                & \textbf{CW-U}                & \textbf{EAD-T}               & \textbf{EAD-U}               & \textbf{Acc}                  \\ \hline
\textbf{ADP}                            & -0.024                       & -0.049                       & -0.087                      & -0.047                       & -0.074                       & -0.12                        & -0.185                       & -0.131                       & -0.012                       & -0.007                       & -0.008                       & -0.006                       & 0.0152                        \\ \hline
\textbf{BaRT-1}                         & 0.015                        & 0.032                        & 0.018                       & 0.029                        & 0.184                        & 0.099                        & 0.089                        & 0.11                         & -0.012                       & -0.057                       & 0.001                        & -0.04                        & -0.0724                       \\ \hline
\textbf{BaRT-10}                        & -0.025                       & 0.006                        & -0.012                      & 0.015                        & -0.142                       & -0.179                       & -0.208                       & -0.182                       & -0.046                       & -0.398                       & -0.053                       & -0.425                       & -0.4384                       \\
\hline
\end{tabular}
\end{adjustbox}
\end{table*}

\begin{table*}
\caption{CIFAR-10 adaptive black-box attack 50\%. Note the defense numbers in the table are the defense accuracy minus the vanilla defense accuracy. This means they are relative accuracies. The very last row is the actual defense accuracy of the vanilla network.}
\centering
\label{table:CIFARM50}
\begin{adjustbox}{width=1\textwidth}
\begin{tabular}{|c|c|c|c|c|c|c|c|c|c|c|c|c|c|}
\hline
                                        & \textbf{FGSM-T}             & \textbf{IFGSM-T}             & \textbf{MIM-T}               & \textbf{PGD-T}               & \textbf{FGSM-U}              & \textbf{IFGSM-U}             & \textbf{MIM-U}               & \textbf{PGD-U}               & \textbf{CW-T}                & \textbf{CW-U}                & \textbf{EAD-T}              & \textbf{EAD-U}               & \textbf{Acc}                  \\ \hline
\textbf{ADP}                            & -0.036                      & -0.116                       & -0.137                       & -0.107                       & -0.077                       & -0.21                        & -0.199                       & -0.229                       & -0.002                       & 0.001                        & 0.006                       & 0.001                        & 0.0152                        \\ \hline
\textbf{BaRT-1}                         & 0.034                       & 0.011                        & 0.021                        & 0.028                        & 0.148                        & 0.071                        & 0.051                        & 0.071                        & -0.009                       & -0.062                       & -0.013                      & -0.064                       & -0.0753                       \\ \hline
\textbf{BaRT-10}                        & 0.036                       & 0.046                        & 0.086                        & 0.037                        & -0.044                       & -0.092                       & -0.008                       & -0.104                       & -0.043                       & -0.414                       & -0.034                      & -0.433                       & -0.4399                       \\
\hline
\end{tabular}
\end{adjustbox}
\end{table*}

\begin{table*}
\caption{CIFAR-10 adaptive black-box attack 75\%. Note the defense numbers in the table are the defense accuracy minus the vanilla defense accuracy. This means they are relative accuracies. The very last row is the actual defense accuracy of the vanilla network.}
\centering
\label{table:CIFARM75}
\begin{adjustbox}{width=1\textwidth}
\begin{tabular}{|c|c|c|c|c|c|c|c|c|c|c|c|c|c|}
\hline
                                        & \textbf{FGSM-T}              & \textbf{IFGSM-T}             & \textbf{MIM-T}               & \textbf{PGD-T}               & \textbf{FGSM-U}              & \textbf{IFGSM-U}             & \textbf{MIM-U}               & \textbf{PGD-U}               & \textbf{CW-T}                & \textbf{CW-U}                & \textbf{EAD-T}               & \textbf{EAD-U}               & \textbf{Acc}                  \\ \hline
\textbf{ADP}                            & -0.038                       & -0.167                       & -0.201                       & -0.167                       & -0.092                       & -0.231                       & -0.216                       & -0.234                       & -0.009                       & -0.006                       & -0.001                       & 0                            & 0.0152                        \\ \hline
\textbf{BaRT-1}                         & 0.002                        & 0.027                        & 0.007                        & 0.015                        & 0.117                        & 0.072                        & 0.029                        & 0.069                        & -0.013                       & -0.067                       & -0.006                       & -0.069                       & -0.0706                       \\ \hline
\textbf{BaRT-10}                        & 0.018                        & 0.052                        & 0.061                        & 0.03                         & -0.055                       & -0.036                       & -0.001                       & -0.03                        & -0.065                       & -0.428                       & -0.058                       & -0.417                       & -0.4349                       \\
\hline
\end{tabular}
\end{adjustbox}
\end{table*}

\begin{table*}
\caption{CIFAR-10 adaptive black-box attack 100\%.Note the defense numbers in the table are the defense accuracy minus the vanilla defense accuracy. This means they are relative accuracies. The very last row is the actual defense accuracy of the vanilla network.}
\centering
\label{table:CIFARM100}
\begin{adjustbox}{width=1\textwidth}
\begin{tabular}{|c|c|c|c|c|c|c|c|c|c|c|c|c|c|}
\hline
                                        & \textbf{FGSM-T}               & \textbf{IFGSM-T}              & \textbf{MIM-T}                & \textbf{PGD-T}                & \textbf{FGSM-U}               & \textbf{IFGSM-U}              & \textbf{MIM-U}                & \textbf{PGD-U}                & \textbf{CW-T}                 & \textbf{CW-U}                 & \textbf{EAD-T}                & \textbf{EAD-U}                & \textbf{Acc}                   \\ \hline
\textbf{ADP}                            & {\color[HTML]{000000} -0.023} & {\color[HTML]{000000} -0.163} & {\color[HTML]{000000} -0.172} & {\color[HTML]{000000} -0.136} & {\color[HTML]{000000} -0.002} & {\color[HTML]{000000} -0.163} & {\color[HTML]{000000} -0.112} & {\color[HTML]{000000} -0.148} & {\color[HTML]{000000} 0.004}  & {\color[HTML]{000000} 0.006}  & {\color[HTML]{000000} -0.004} & {\color[HTML]{000000} -0.002} & {\color[HTML]{000000} 0.0152}  \\ \hline
\textbf{BaRT-1}                         & {\color[HTML]{000000} 0.02}   & {\color[HTML]{000000} 0.011}  & {\color[HTML]{000000} 0.022}  & {\color[HTML]{000000} 0.027}  & {\color[HTML]{000000} 0.173}  & {\color[HTML]{000000} 0.166}  & {\color[HTML]{000000} 0.139}  & {\color[HTML]{000000} 0.169}  & {\color[HTML]{000000} -0.016} & {\color[HTML]{000000} -0.069} & {\color[HTML]{000000} -0.018} & {\color[HTML]{000000} -0.054} & {\color[HTML]{000000} -0.0707} \\ \hline
\textbf{BaRT-10}                        & {\color[HTML]{000000} 0.044}  & {\color[HTML]{000000} 0.05}   & {\color[HTML]{000000} 0.126}  & {\color[HTML]{000000} 0.073}  & {\color[HTML]{000000} 0.057}  & {\color[HTML]{000000} 0.078}  & {\color[HTML]{000000} 0.123}  & {\color[HTML]{000000} 0.118}  & {\color[HTML]{000000} -0.063} & {\color[HTML]{000000} -0.405} & {\color[HTML]{000000} -0.047} & {\color[HTML]{000000} -0.404} & {\color[HTML]{000000} -0.4409} \\
\hline
\end{tabular}
\end{adjustbox}
\end{table*}

\begin{table*}
\caption{Fashion-MNIST pure black-box attack. Note the defense numbers in the table are the defense accuracy minus the vanilla defense accuracy. This means they are relative accuracies. The very last row is the actual defense accuracy of the vanilla network.}
\centering
\label{table:FMPure}
\begin{adjustbox}{width=1\textwidth}
\begin{tabular}{|c|c|c|c|c|c|c|c|c|c|c|c|c|c|}
\hline
{\color[HTML]{000000} }                 & {\color[HTML]{000000} \textbf{FGSM-T}} & {\color[HTML]{000000} \textbf{IFGSM-T}} & {\color[HTML]{000000} \textbf{MIM-T}} & {\color[HTML]{000000} \textbf{PGD-T}} & {\color[HTML]{000000} \textbf{CW-T}} & {\color[HTML]{000000} \textbf{EAD-T}} & {\color[HTML]{000000} \textbf{FGSM-U}} & {\color[HTML]{000000} \textbf{IFGSM-U}} & {\color[HTML]{000000} \textbf{MIM-U}} & {\color[HTML]{000000} \textbf{PGD-U}} & {\color[HTML]{000000} \textbf{CW-U}} & {\color[HTML]{000000} \textbf{EAD-U}} & {\color[HTML]{000000} \textbf{Acc}} \\ \hline
{\color[HTML]{000000} \textbf{ADP}}     & {\color[HTML]{000000} 0.01}            & {\color[HTML]{000000} -0.05}            & {\color[HTML]{000000} -0.035}         & {\color[HTML]{000000} -0.018}         & {\color[HTML]{000000} -0.003}        & {\color[HTML]{000000} -0.002}         & {\color[HTML]{000000} -0.031}          & {\color[HTML]{000000} -0.006}           & {\color[HTML]{000000} -0.03}          & {\color[HTML]{000000} -0.024}         & {\color[HTML]{000000} 0.012}         & {\color[HTML]{000000} 0.018}          & {\color[HTML]{000000} 0.013}        \\ \hline
{\color[HTML]{000000} \textbf{BaRT-1}}  & {\color[HTML]{000000} 0.041}           & {\color[HTML]{000000} 0.055}            & {\color[HTML]{000000} 0.089}          & {\color[HTML]{000000} 0.062}          & {\color[HTML]{000000} -0.002}        & {\color[HTML]{000000} 0}              & {\color[HTML]{000000} 0.119}           & {\color[HTML]{000000} 0.173}            & {\color[HTML]{000000} 0.137}          & {\color[HTML]{000000} 0.159}          & {\color[HTML]{000000} -0.021}        & {\color[HTML]{000000} -0.017}         & {\color[HTML]{000000} -0.0317}      \\
\hline
\end{tabular}
\end{adjustbox}
\end{table*}

\begin{table*}
\caption{Fashion-MNIST adaptive black-box attack 1\%. Note the defense numbers in the table are the defense accuracy minus the vanilla defense accuracy. This means they are relative accuracies. The very last row is the actual defense accuracy of the vanilla network.}
\centering
\label{table:FM1}
\begin{adjustbox}{width=1\textwidth}
\begin{tabular}{|c|c|c|c|c|c|c|c|c|c|c|c|c|c|}
\hline
                                        & \textbf{FGSM-T}              & \textbf{IFGSM-T}             & \textbf{MIM-T}               & \textbf{PGD-T}               & \textbf{FGSM-U}             & \textbf{IFGSM-U}             & \textbf{MIM-U}               & \textbf{PGD-U}               & \textbf{CW-T}                & \textbf{CW-U}               & \textbf{EAD-T}               & \textbf{EAD-U}               & \textbf{Acc}                  \\ \hline
\textbf{ADP}                            & 0.03                         & -0.018                       & -0.009                       & -0.004                       & 0.051                       & -0.029                       & 0.008                        & -0.027                       & -0.018                       & 0.023                       & 0.005                        & 0.031                        & 0.013                         \\ \hline
\textbf{BaRT-1}                         & 0.083                        & 0.063                        & 0.05                         & 0.061                        & 0.229                       & 0.137                        & 0.175                        & 0.171                        & 0.041                        & -0.022                      & 0.009                        & -0.026                       & -0.0308                       \\ \hline
\textbf{BaRT-4}                         & 0.069                        & 0.04                         & 0.034                        & 0.049                        & 0.153                       & 0.056                        & 0.067                        & 0.055                        & -0.035                       & -0.182                      & -0.032                       & -0.165                       & -0.0999                       \\
\hline
\end{tabular}
\end{adjustbox}
\end{table*}

\begin{table*}
\caption{Fashion-MNIST adaptive black-box attack 25\%. Note the defense numbers in the table are the defense accuracy minus the vanilla defense accuracy. This means they are relative accuracies. The very last row is the actual defense accuracy of the vanilla network.}
\centering
\label{table:FM25}
\begin{adjustbox}{width=1\textwidth}
\begin{tabular}{|c|c|c|c|c|c|c|c|c|c|c|c|c|c|}
\hline
                                        & \textbf{FGSM-T}              & \textbf{IFGSM-T}            & \textbf{MIM-T}               & \textbf{PGD-T}               & \textbf{FGSM-U}              & \textbf{IFGSM-U}             & \textbf{MIM-U}               & \textbf{PGD-U}               & \textbf{CW-T}                & \textbf{CW-U}                & \textbf{EAD-T}               & \textbf{EAD-U}               & \textbf{Acc}                  \\ \hline
\textbf{ADP}                            & 0.089                        & 0.025                       & 0.036                        & 0.004                        & 0.035                        & -0.038                       & -0.008                       & -0.038                       & -0.04                        & -0.092                       & -0.013                       & -0.032                       & 0.013                         \\ \hline
\textbf{BaRT-1}                         & 0.11                         & 0.238                       & 0.195                        & 0.217                        & 0.191                        & 0.165                        & 0.159                        & 0.145                        & -0.038                       & -0.134                       & -0.019                       & -0.097                       & -0.0314                       \\ \hline
\textbf{BaRT-4}                         & 0.141                        & 0.293                       & 0.268                        & 0.256                        & 0.215                        & 0.246                        & 0.224                        & 0.256                        & -0.067                       & -0.216                       & -0.041                       & -0.218                       & -0.1018                       \\
\hline
\end{tabular}
\end{adjustbox}
\end{table*}

\begin{table*}
\caption{Fashion-MNIST adaptive black-box attack 50\%. Note the defense numbers in the table are the defense accuracy minus the vanilla defense accuracy. This means they are relative accuracies. The very last row is the actual defense accuracy of the vanilla network.}
\centering
\label{table:FM50}
\begin{adjustbox}{width=1\textwidth}
\begin{tabular}{|c|c|c|c|c|c|c|c|c|c|c|c|c|c|}
\hline
                                        & \textbf{FGSM-T}              & \textbf{IFGSM-T}             & \textbf{MIM-T}               & \textbf{PGD-T}               & \textbf{FGSM-U}              & \textbf{IFGSM-U}             & \textbf{MIM-U}               & \textbf{PGD-U}               & \textbf{CW-T}                & \textbf{CW-U}                & \textbf{EAD-T}              & \textbf{EAD-U}               & \textbf{Acc}                  \\ \hline
\textbf{ADP}                            & 0.115                        & -0.015                       & 0.021                        & -0.024                       & 0.01                         & -0.022                       & -0.013                       & -0.027                       & -0.019                       & -0.065                       & -0.011                      & -0.044                       & 0.013                         \\ \hline
\textbf{BaRT-1}                         & 0.143                        & 0.227                        & 0.263                        & 0.24                         & 0.183                        & 0.205                        & 0.197                        & 0.195                        & -0.047                       & -0.114                       & -0.02                       & -0.1                         & -0.0312                       \\ \hline
\textbf{BaRT-4}                         & 0.179                        & 0.325                        & 0.327                        & 0.314                        & 0.241                        & 0.246                        & 0.258                        & 0.224                        & -0.059                       & -0.216                       & -0.024                      & -0.184                       & -0.1                          \\
\hline
\end{tabular}
\end{adjustbox}
\end{table*}

\begin{table*}
\caption{Fashion-MNIST adaptive black-box attack 75\%. Note the defense numbers in the table are the defense accuracy minus the vanilla defense accuracy. This means they are relative accuracies. The very last row is the actual defense accuracy of the vanilla network.}
\centering
\label{table:FM75}
\begin{adjustbox}{width=1\textwidth}
\begin{tabular}{|c|c|c|c|c|c|c|c|c|c|c|c|c|c|}
\hline
                                        & \textbf{FGSM-T}              & \textbf{IFGSM-T}             & \textbf{MIM-T}               & \textbf{PGD-T}              & \textbf{FGSM-U}              & \textbf{IFGSM-U}             & \textbf{MIM-U}               & \textbf{PGD-U}               & \textbf{CW-T}               & \textbf{CW-U}                & \textbf{EAD-T}               & \textbf{EAD-U}               & \textbf{Acc}                  \\ \hline
\textbf{ADP}                            & 0.089                        & 0.024                        & 0.103                        & 0.018                       & -0.033                       & -0.028                       & -0.026                       & -0.016                       & -0.016                      & -0.054                       & -0.006                       & -0.054                       & 0.013                         \\ \hline
\textbf{BaRT-1}                         & 0.144                        & 0.299                        & 0.336                        & 0.297                       & 0.151                        & 0.239                        & 0.195                        & 0.254                        & -0.027                      & -0.112                       & -0.002                       & -0.078                       & -0.0316                       \\ \hline
\textbf{BaRT-4}                         & 0.173                        & 0.347                        & 0.41                         & 0.345                       & 0.196                        & 0.304                        & 0.269                        & 0.36                         & -0.046                      & -0.17                        & -0.022                       & -0.167                       & -0.107                        \\
\hline
\end{tabular}
\end{adjustbox}
\end{table*}

\begin{table*}[t]
\caption{Fashion-MNIST adaptive black-box attack 100\%. Note the defense numbers in the table are the defense accuracy minus the vanilla defense accuracy. This means they are relative accuracies. The very last row is the actual defense accuracy of the vanilla network.}
\centering
\label{table:FM100}
\begin{adjustbox}{width=1\textwidth}
\begin{tabular}{|c|c|c|c|c|c|c|c|c|c|c|c|c|c|}
\hline
                                        & \textbf{FGSM-T}              & \textbf{IFGSM-T}             & \textbf{MIM-T}              & \textbf{PGD-T}               & \textbf{FGSM-U}              & \textbf{IFGSM-U}             & \textbf{MIM-U}               & \textbf{PGD-U}               & \textbf{CW-T}                & \textbf{CW-U}                & \textbf{EAD-T}               & \textbf{EAD-U}               & \textbf{Acc}                  \\ \hline
\textbf{ADP}                            & 0.086                        & -0.014                       & 0.039                       & -0.012                       & -0.093                       & -0.038                       & -0.007                       & -0.029                       & -0.006                       & -0.027                       & -0.005                       & -0.03                        & 0.013                         \\ \hline
\textbf{BaRT-1}                         & 0.129                        & 0.278                        & 0.304                       & 0.274                        & 0.125                        & 0.26                         & 0.228                        & 0.258                        & -0.015                       & -0.1                         & -0.015                       & -0.062                       & -0.0317                       \\ \hline
\textbf{BaRT-4}                         & 0.165                        & 0.319                        & 0.383                       & 0.317                        & 0.176                        & 0.276                        & 0.273                        & 0.288                        & -0.052                       & -0.182                       & -0.032                       & -0.148                       & -0.1062                       \\
\hline
\end{tabular}
\end{adjustbox}
\end{table*}

\begin{table*}
\caption{Clean prediction accuracy of the Odds defense on Fashion-MNIST and CIFAR-10 with different FPRs.}
\centering
\begin{tabular}{|l|l|l|l|l|l|l|l|}
\hline
FPR             & 1\%   & 10\%  & 20\%  & 30\% & 40\% & 50\% & 80\%   \\ \hline 
FashionMNIST    & 78.6\%  & 79.6\%  & 78.5\%  & 79.5\% & 78.6\% & 78.8\% & 79.1\%   \\ \hline 
CIFAR-10         &  0.3\%  & 27.8\%  & 43.2\%  & 61.1\% & 75.2\% & 86.2\% & 99.3\%   \\ \hline
\end{tabular}
\label{tab:clean_accuracy_odds_odd_cifar_fashionmnist}
\end{table*}

\end{document}